\documentclass[10pt, a4paper]{article}

\usepackage[utf8]{inputenc}
\usepackage[T1]{fontenc}
\usepackage[english]{babel}
\usepackage{geometry}
\usepackage{booktabs} 
\usepackage{graphicx} 
\usepackage{xcolor}
\usepackage{fancyhdr}
\usepackage{authblk} 

\definecolor{bilipink}{RGB}{251, 114, 153}
\definecolor{biliblue}{RGB}{0, 140, 210}

\usepackage{listings}
\usepackage[ruled,vlined]{algorithm2e}
\definecolor{commentcolor}{RGB}{110,110,110}

\SetCommentSty{mycommfont}

\usepackage[default]{sourcesanspro}

\usepackage[
    colorlinks,
    citecolor=bilipink,
    linkcolor=bilipink,
    urlcolor=bilipink,
    breaklinks=true
]{hyperref}

\usepackage{amsmath}
\usepackage{amsfonts}
\usepackage{multirow}
\usepackage{subcaption}
\usepackage{tcolorbox}
\tcbuselibrary{listings,skins,breakable}
\usepackage{enumitem}

\makeatletter
\def\@maketitle{%
  \newpage
  \null
  \vspace{-3.5em} 
  \begin{center}%
    {\color{biliblue}\hrule height 0.8pt}
    \vspace{1.5em}
    
    {\huge \bfseries \@title \par} 
    \vspace{1.5em}
    
    {\color{biliblue}\hrule height 0.8pt}
    \vspace{1.8em}
    
    {\large \bfseries \@author} \\
    \vspace{0.5em}
  \end{center}%
  \par
  \vspace{1.5em}}
\makeatother

\usepackage{hyperref}
\hypersetup{
    colorlinks=true,
    urlcolor=bilipink, 
}

\title{Community-Aware Assessment of Social Textual Engagement and Resonance: A Human-Centric Perspective on User-Generated Content Evaluation}
\author{\textbf{Tianjiao Li\textsuperscript{*\dag}, Kai Zhao\textsuperscript{*}, Xiang Li, Yang Liu, Huyang Sun}}
\affil{Bilibili Index LLM Team}
\affil{\textsuperscript{*}Equal contribution \quad \textsuperscript{\dag}Corresponding author}
\affil{\small\texttt{\{litianjiao01, zhaokai02, lixiang19, liuyang17, sunhuyang\}@bilibili.com}}
\date{} 

\begin{document}

\maketitle
\thispagestyle{fancy}

\begin{abstract}
Traditional Video Quality Assessment (VQA) focuses narrowly on aesthetic fidelity, overlooking the complex social dynamics that define quality in User-Generated Content (UGC). In this work, we propose a paradigm shift from signal-centric metrics to human-centric resonance assessment. We introduce CASTER (Community-Aware Assessment of Social Textual Engagement and Resonance), a new task that evaluates whether a UGC item achieves positive community resonance based on its multimodal attributes rather than visual quality alone. To address this, we present MEDEA (Multimodal Engagement-Driven Evaluation Architecture), which introduces a novel Social Chain-of-Thought (Social-CoT) mechanism. Unlike traditional logical CoT, Social-CoT performs multimodal perspective-taking, instantiating diverse viewer personas to simulate collective cognitive and emotional reactions (i.e., the "community mind") before deriving a quality judgment. MEDEA is trained via a two-stage approach involving supervised fine-tuning and process-supervised reinforcement learning with Social Alignment Reward to ensure reasoning paths are grounded in authentic human social cognition. To support this task, we release CASTER-Bench, a comprehensive human-annotated benchmark covering diverse UGC categories. Experiments demonstrate that MEDEA significantly outperforms state-of-the-art baselines on CASTER-Bench while providing interpretable and empathetic reasoning paths that align with real community feedback.
\end{abstract}

\begin{center}
\def\huggingface{\raisebox{-1.5pt}{\includegraphics[height=1.05em]{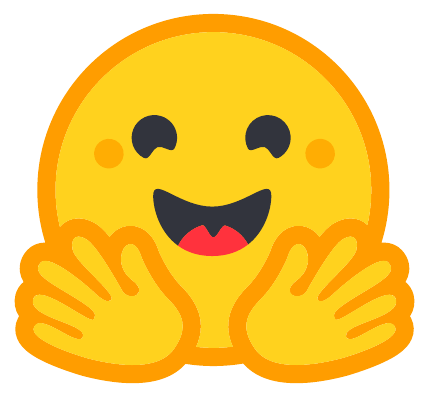}}}
\def\github{\raisebox{-1.5pt}{\includegraphics[height=1.05em]{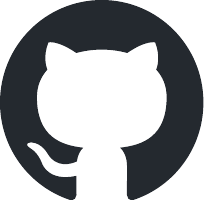}}}
\begin{tabular}{cl}
\github & \url{https://github.com/bilibili/medea_rl}\\
\huggingface & \url{https://huggingface.co/IndexTeam/MEDEA}\\
\huggingface & \url{https://huggingface.co/datasets/IndexTeam/CASTER-Bench}\\
\end{tabular}
\end{center}

\section{Introduction}

Traditional Video Quality Assessment (VQA) has achieved notable success in measuring aesthetic fidelity and technical distortions~\cite{seshadrinathan2010study, lin2015mcl, danier2023bvi}. However, its core objective is fundamentally misaligned with how quality is perceived on User-Generated Content (UGC) platforms. By focusing primarily on pixel level integrity and low-level visual cues, existing VQA methods~\cite{wu2023exploring, wu2023towards, lu2024kvq, duan2025finevq} fail to capture the human-centered and social nature of quality in real-world UGC. As a result, these approaches struggle to reflect whether content is meaningful, engaging, or valuable to actual users beyond momentary visual appeal.

The key challenge, therefore, lies in how to properly define UGC quality. On large-scale platforms, high-quality content is determined not by technical perfection, but by whether it resonates with the community eliciting emotional engagement, meaningful discussion, and positive recognition. Such community endorsement is most explicitly reflected through user engagement signals, among which positive comments provide direct, content level evidence of perceived quality.

While Large Language Models (LLMs) have demonstrated impressive reasoning capabilities via Chain-of-Thought (CoT) in logical and mathematical domains~\cite{wei2022chain}, \textit{Social Reasoning}, the ability to model human emotional dynamics and collective reception remains underexplored. We argue that assessing UGC quality requires a Theory of Mind (ToM) approach~\cite{sap2022neural}: the model must not merely analyze the content signals, but actively ``step into the shoes'' of the audience. We term this process \textit{Social Chain-of-Thought (Social-CoT)}, where the model explicitly generates a diverse set of empathetic reaction paths simulating the ``community mind'' before converging on a quality judgment.

Motivated by this observation, we introduce \textit{CASTER} (Community-Aware Assessment of Social Textual Engagement and Resonance), a task that reframes UGC quality assessment as identifying content genuinely endorsed by its audience via social reasoning.

However, direct access to user comments is often unavailable, especially for newly uploaded or sparsely interacted content, where quality assessment is still critically needed for recommendation and moderation. To address this limitation, we propose \textit{MEDEA} (Multimodal Engagement-Driven Evaluation Architecture), which operationalizes the Social-CoT paradigm. MEDEA infers community resonance by instantiating diverse viewer personas and simulating plausible user comments conditioned on multimodal content signals, effectively performing multimodal perspective-taking before aggregating these reaction paths into a final quality judgment.

To achieve this capability, MEDEA is trained via supervised fine-tuning (SFT) and process-supervised reinforcement learning (RL), combining large-scale pseudo-labeled data with expert annotations. Crucially, we introduce \textit{Social Alignment Reward} during the RL stage to ensure the generated reasoning paths are grounded in authentic human social cognition rather than robotic analysis. Experiments demonstrate that MEDEA substantially outperforms aesthetic and multimodal baselines~\cite{wu2022fastquality, wu2023towards, wu2024q, duan2025finevq, jia2025vqa2}, while providing interpretable and community-aligned reasoning traces.

Furthermore, to support this task, we present \emph{CASTER-Bench}, a multimodal benchmark specifically designed for long-form UGC videos, with an average duration of 442 seconds. Unlike existing VQA datasets that predominantly rely on short clips (typically 8-10 seconds), CASTER-Bench enables the evaluation of narrative coherence, information density, and sustained engagement that are critical in real-world content recommendation scenarios. The benchmark is annotated by expert raters using a human-centered rubric, and empirical analysis reveals a strong colorrelation between positive user comments and expert judgments, while traditional VQA and vision-centric models perform poorly. These results highlight the limitations of existing methods in modeling the semantic, social, and temporal factors underlying UGC quality.

Our contributions are summarized as follows:
\begin{itemize}[nosep]
    \item We introduce  CASTER, a community-aware task that redefines UGC quality through the lens of social reasoning, and release CASTER-Bench, a multimodal benchmark annotated using a human-centered rubric.
    \item We propose MEDEA, an evaluation framework that pioneers Social-CoT to simulate empathetic user reactions, trained via SFT and process-supervised RL with Social Alignment Reward.
    \item We demonstrate that MEDEA significantly outperforms diverse types of baselines while offering improved interpretability through generated social reasoning paths.
\end{itemize}

\section{Related Works}
\label{sec:related_works}

\subsection{UGC Databases}
Early UGC benchmarks \cite{seshadrinathan2010study, lin2015mcl, nuutinen2016cvd2014, mackin2018study, madhusudana2021subjective, danier2023bvi} mainly relied on professionally produced videos with controlled, synthetic distortions. Recent datasets have shifted focus toward authentic, in-the-wild UGC with large-scale crowdsourced annotations, including KoNViD-1k \cite{hosu2017konstanz}, LIVE-VQC \cite{sinno2018large}, YouTube-UGC \cite{wang2019youtube}, and PUGCQ \cite{li2021pugcq}, which better reflect real-world content diversity and mixed distortions. 

Beyond overall quality scores, recent efforts have moved toward multi-dimensional quality modeling by disentangling aesthetic and technical factors. Notable examples include datasets explored in DOVER~\cite{wu2023exploring}, MD-VQA~\cite{zhang2023md}, MaxVQA~\cite{wu2023towards}, KVQ~\cite{lu2024kvq}, and FineVQ~\cite{duan2025finevq}. In parallel, VF-EVAL~\cite{song2025vf} introduces a benchmark for evaluating MLLMs’ ability to generate feedback on AIGC videos, focusing on prompt alignment, coherence, and commonsense reasoning. However, these datasets and benchmarks predominantly emphasize perceptual attributes or feedback correctness for short-form or synthetic videos. In contrast, CASTER-Bench targets long-form, real-world UGC and explicitly models social-cognitive judgments such as narrative engagement and emotional resonance, which are critical for understanding community-level content appreciation.

\subsection{UGC-VQA Models}
UGC-VQA methods have evolved from full-reference metrics \cite{manasa2016optical, wang2007video, moorthy2010efficient, madhusudana2021st, vu2011spatiotemporal}, which require unavailable references, to no-reference approaches. Classical models leveraged handcrafted statistical priors \cite{wang2002no}, while modern approaches learn content-dependent spatiotemporal representations from large-scale distorted data \cite{varga2022no, de2024no, li2019quality, zhang2023md, xing2022starvqa, ge2025lmm, duan2025finevq}. Representative methods include VSFA \cite{li2019quality} (temporal modeling), MD-VQA \cite{zhang2023md} (fusion of spatial, motion, and semantic cues), StarVQA \cite{xing2022starvqa} (self-attention on salient spatiotemporal regions), and DOVER \cite{wu2023exploring} (dual-branch modeling of technical quality and aesthetic preference).

The recent advent of vision-language pretraining has catalyzed multimodal directions in UGC-VQA \cite{radford2021learning, tschannen2025siglip}. CLIP-based methods, such as COVER \cite{he2024cover} and MaxVQA \cite{wu2023towards}, employ semantic encoders to inject high-level content priors. Furthermore, prompt-driven alignment methods like Q-Align \cite{wu2024q} enable zero-shot or cross-modal approximation of human judgments. Emerging Large Multimodal Models (LMMs), such as LMM-VQA \cite{ge2025lmm}, FineVQ \cite{duan2025finevq}, and CAMP-VQA \cite{wang2025camp}, integrate spatial, temporal, and text-based reasoning to produce robust quality estimates. However, these methods typically treat text as a static feature rather than utilizing it to simulate the dynamic social reception of the content.

\subsection{Chain-of-Thought and Social Intelligence}
\label{sec:related_social_cot}
While Chain-of-Thought (CoT) prompting has revolutionized large language model performance in logical, mathematical, and symbolic reasoning tasks \cite{wei2022chain, shao2024deepseekmathpushinglimitsmathematical}, its application to social intelligence remains a frontier challenge. Recent studies in Theory of Mind (ToM) investigate whether LLMs can effectively infer the mental states, beliefs, and emotional reactions of others \cite{sap2022neural}. In the context of UGC assessment, we argue that quality is not an intrinsic property of the signal but a product of social reception.

Our work bridges these domains by proposing Social-CoT. Unlike standard CoT which focuses on step-by-step logical deduction, Social-CoT explicitly operationalizes ToM by simulating diverse viewer personas and their empathetic engagement paths. This approach shifts the evaluation paradigm from analyzing static content features to simulating the ``community mind'', thereby aligning computational quality assessment with authentic community dynamics.

\section{Community-Aware Assessment of Social Textual Engagement and Resonance}

This section formalizes the CASTER task and introduces CASTER-Bench, a benchmark designed to support this task.
We describe the UGC item collection process, expert-driven annotation protocol, and quality control procedures, followed by dataset statistics and comparisons with existing benchmarks.

\subsection{The CASTER Task}

CASTER aims to assess whether a piece of user-generated content resonates with the community from a holistic, human-centric perspective.
Unlike traditional video quality assessment which focuses on low-level aesthetic or technical attributes (e.g., sharpness or noise), CASTER evaluates the quality of the content artifact itself rather than the video signal alone.

Formally, given a UGC item consisting of multimodal inputs including video frames, cover image, title, tags, category metadata, and automatic speech recognition (ASR) transcripts, the task is to predict whether the content is perceived as \emph{high-quality} or \emph{low-quality} according to human judgment.
This judgment reflects community-level resonance and is shaped by factors such as creativity, emotional engagement, informational value, narrative coherence, and originality. By framing quality assessment as a community-aware and content-driven task, CASTER decouples perceived quality from confounding signals such as view count or recommendation exposure, better aligning automatic evaluation with real user preferences.

\begin{figure*}[t]
    \centering
    \begin{subfigure}[t]{0.695\textwidth}
        \centering
        \includegraphics[width=\linewidth]{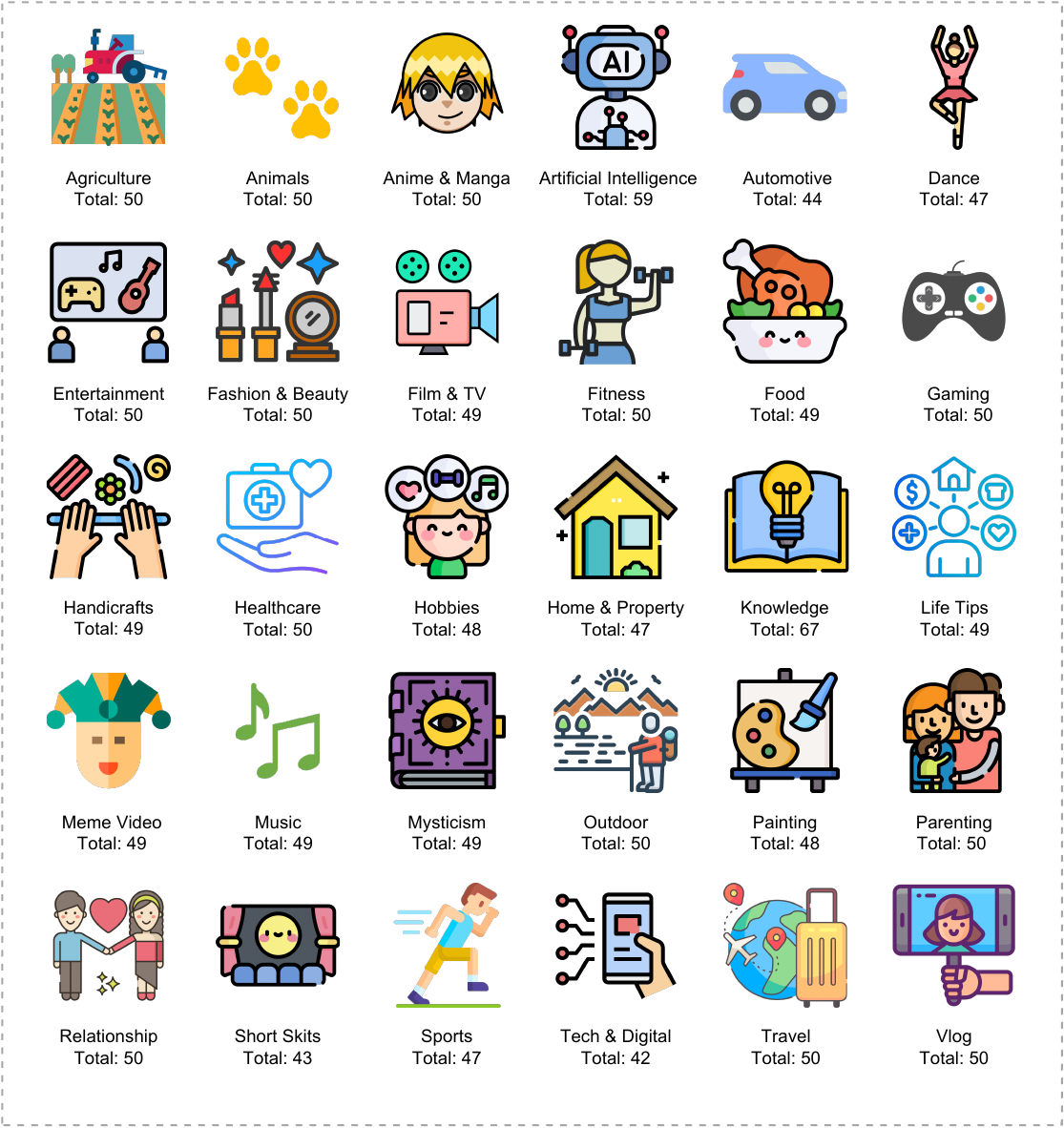}
        \caption{Category-level distribution of CASTER-Bench across 30 major UG categories.}
        \label{fig:category30_a}
    \end{subfigure}
    \hfil
    \begin{subfigure}[t]{0.29\textwidth}
        \centering
        \includegraphics[width=\linewidth]{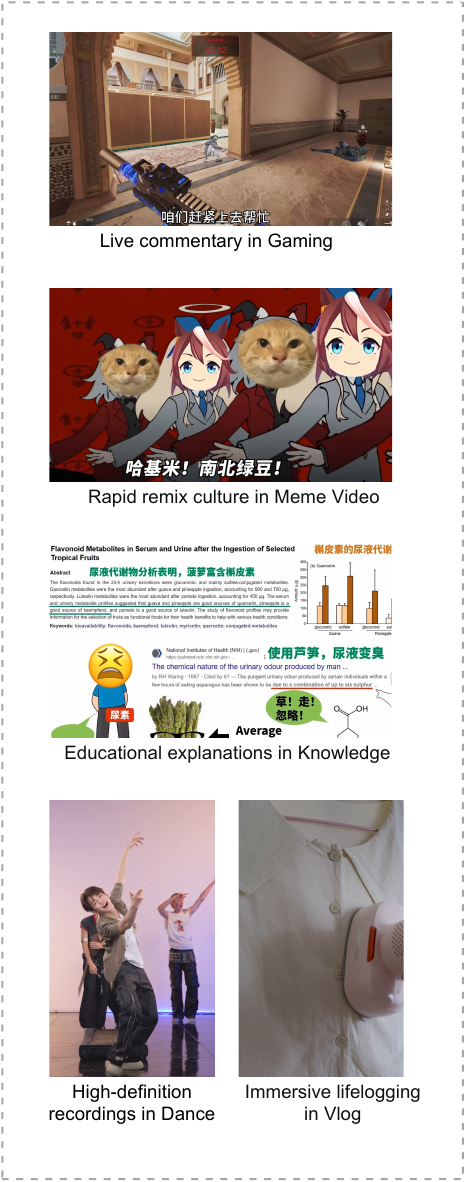}
        \caption{Representative UGC examples.}
        \label{fig:category30_b}
    \end{subfigure}

    \caption{Overview of CASTER-Bench. (a) Category-level composition of the benchmark, covering 1,485 UGC items sampled from 30 major content categories with balanced representation. (b) Representative examples illustrating diverse presentation styles and content paradigms, including live commentary, meme remix culture, educational explanations, high-definition performance recordings, and immersive vlogging.}
    \label{fig_30categories}
\end{figure*}

\subsection{CASTER-Bench: A Benchmark for Social Resonance}
To support the CASTER task, we introduce CASTER-Bench, a human-annotated benchmark containing 1,485 UGC items curated from a large-scale comprehensive video platform and annotated by professional content operation experts.

In contrast to existing benchmarks such as KVQ~\cite{lu2024kvq} and FineVD~\cite{duan2025finevq}, which emphasize aesthetic quality on short clips, CASTER-Bench focuses on subjective, multidimensional perceptions of long-form content quality (average 442s), including creativity, emotional value, informational utility, and narrative excellence. Each item is accompanied by rich multimodal information, including visual content, cover image, title, tags, category metadata, and ASR transcripts, enabling holistic assessment beyond visual appearance alone.

\subsubsection{Data Collection and Statistics}
UGC items were collected following stratified random sampling across 30 major content categories (e.g., \textit{Lifestyle}, \textit{Knowledge}, \textit{Gaming}) to ensure broad coverage of diverse content scenarios, as illustrated in Figure ~\ref{fig:category30_a}. Figure ~\ref{fig:category30_b} also demonstrates representative examples, highlighting the diversity in content forms and production paradigms.

CASTER-Bench contains 1,485 UGC items with a quality label distribution mirroring real-world platforms: \textit{Excellent} (10.6\%), \textit{Good} (17.0\%), \textit{Average} (38.6\%), and \textit{Poor} (33.7\%). This distribution presents a realistic challenge for identifying high-quality content amidst massive amounts of average data.

\begin{table*}[htbp]
\centering
\footnotesize
\begin{tabular*}{\textwidth}{@{\extracolsep{\fill}}lccccccc@{}}
\toprule
\textbf{Database} & \textbf{Source} & \textbf{Num.} & \textbf{Avg Dur.} & \textbf{Total Dur.} & \textbf{Focus} & \textbf{Modality}& \textbf{Env.}\\
\midrule
KoNViD-1k~\cite{hosu2017konstanz} & YFCC100m & 1200 & 8s & 2.7h & A\&T & Video Only & Crowd \\
LIVE-VQC~\cite{sinno2018large} & UGC-P & 585 & 10s & 1.6h & A\&T & Video Only & Crowd \\
YouTube-UGC~\cite{wang2019youtube} & UGC-P & 1380 & 20s & 7.7h & A\&T & Video Only & Crowd \\
KVQ~\cite{lu2024kvq} & UGC-P & 600 & 8s & 8h & A\&T & Video Only & In-lab \\
FineVD~\cite{duan2025finevq} & UGC-P & 6104 & 8s & 13.6h & A\&T & Video Only & In-lab \\
\textbf{CASTER-Bench (Ours)} & UGC-P & 1485 & \textbf{442s} & \textbf{182.5h} & \textbf{S\&C} & \textbf{T\&T\&V\&A} & In-lab \\
\bottomrule
\end{tabular*}
\caption{Multi-dimension comparison between mainstream general video quality assessment datasets. 
Num. denotes the total number of test video sequences;
Avg Dur. and Total Dur. denote average duration per video (seconds) and combined duration of all videos (hours).
A\&T and S\&C indicate aesthetic–technical and subjective content-driven quality;
T\&T\&V\&A includes title, tags, video, and ASR transcripts;
Crowd and In-lab denote annotation environments.}
\label{table_comparison}
\normalsize 
\end{table*}

\subsubsection{Expert-Driven Annotation Protocol}
To ensure the reliability, consistency, and practical relevance of the annotations, we adopt a rigorously designed expert-driven annotation protocol grounded in real-world content moderation and recommendation practices.
In particular, we recruited 10 professional content operation experts to annotate the dataset. The annotation is based on a comprehensive framework comprising four core dimensions:
\begin{itemize}[nosep]
    \item \textbf{Production Quality}: audiovisual execution, post-production, and special effects.
    \item \textbf{Perceived Value}: emotional resonance, entertainment, or affective engagement.
    \item \textbf{Information Utility}: practical knowledge, instructional value, or curated information.
    \item \textbf{Narrative Excellence}: coherent structure, originality, or innovative presentation.
\end{itemize}

Annotators labeled items as \emph{Excellent}, \emph{Good}, \emph{Average}, or \emph{Poor}. Crucially, they received high-engagement user comments and were instructed to use them as complementary evidence to judge whether content elicited genuine community resonance, rather than relying solely on visual signals.

A core objective of CASTER-Bench is to assess the \textit{intrinsic value} of UGC rather than merely predicting popularity metrics like view counts, which are often saturated with noise such as recommendation biases and sensationalist tactics. The expert annotations serve as a ``refinement'' mechanism, filtering out confounding factors to prioritize genuine community resonance over superficial traffic. Detailed case studies distinguishing high-popularity content from high-quality content are provided in Appendix~\ref{app:intrinsic_vs_popularity}. A sanitized version of the data will be provided in the final camera-ready version.

\section{Multimodal Engagement-Driven Evaluation Architecture}
\label{sec:method}

\begin{figure*}[t]
\centering
  \includegraphics[width=6.35in]{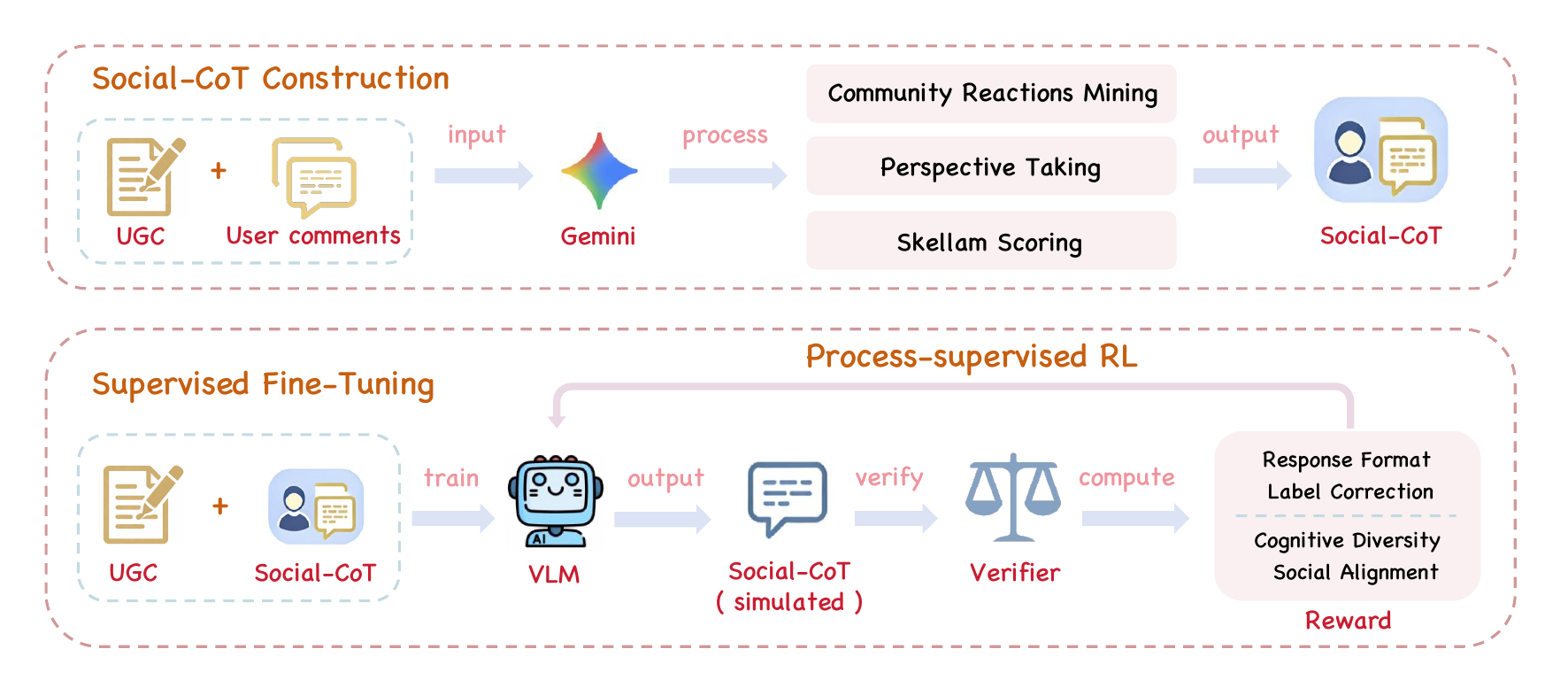}
  \caption{Overview of the MEDEA framework. The upper part depicts the Social-CoT construction pipeline, including community reactions mining, perspective taking, and consensus mechanism via Skellam Scoring. The lower part illustrates the training procedure, consisting of supervised fine-tuning and process-supervised reinforcement learning with multiple reward signals.}
 \label{fig_structure}
\end{figure*}

In this section, we propose MEDEA, a unified framework that operationalizes the Social-CoT paradigm. Rather than mapping multimodal signals directly to a quality label, MEDEA simulates a ``community of critics'' by generating diverse empathetic reasoning paths before aggregating them into a final judgment. MEDEA follows a three-stage pipeline: (1) constructing a large-scale Social-CoT corpus by mining community reactions and instantiating viewer personas; (2) supervised fine-tuning to internalize the capability of multimodal perspective-taking; and (3) process-supervised reinforcement learning with Social Alignment Reward to refine the authenticity and diversity of the social reasoning process. Figure~\ref{fig_structure} provides an overview of the MEDEA framework.

\subsection{Constructing Social-CoT Paths}
\label{sec:data_construction}

To train a model capable of social reasoning, we construct a dataset that transforms raw UGC engagement signals into structured empathetic reasoning paths. We combine large-scale unlabeled scripts containing real user comments with a smaller, expert-annotated dataset.

\paragraph{Mining Community Reactions and Perspective Taking.}
We posit that understanding UGC quality requires identifying specific ``viewer personas'' within the community. Given a UGC item, we treat its comment section as a reflection of the collective ``community mind''. For unlabeled data, we retrieve the top-50 most-liked comments and employ a teacher model to filter for relevance, selecting 15-20 reactions that capture core dimensions such as creativity, emotional appeal, and narrative structure. These selected comments serve as authentic reaction anchors. For the reasoning process, we instruct Gemini-2.5-Flash to perform multimodal perspective-taking: it must instantiate diverse viewer personas and articulate why specific visual or narrative elements trigger specific reactions (refer to Appendix~\ref{app:prompts} for the detailed prompts). For data with expert-provided labels, we apply the same prompting pipeline but explicitly instruct the teacher model to ensure that both its reasoning process and final answer agree with the gold label.

\paragraph{Consensus Mechanism via Skellam Scoring.}
To transit from diverse social perspectives to a unified quality judgment, we implement a statistical consensus mechanism. Each reasoning path (simulated comment) is assigned a supportive or oppositional stance. Let $X$ denote the number of supportive paths and $Y$ denote the number of oppositional paths. We compute the Skellam-normalized difference score $z$ to model the significance of the community endorsement:
\begin{equation}
    z = \frac{X - Y}{\sqrt{X + Y}}.
\end{equation}
A heuristic quality label is then assigned based on this community consensus:
\begin{equation}
    \text{label} = 
    \begin{cases} 
    \text{High-Quality}, & \text{if } z \ge 1.5, \\
    \text{Low-Quality}, & \text{otherwise.}
    \end{cases}
\end{equation}
This ``Think-then-Aggregate'' structure forms the training target for our Social-CoT, ensuring the final judgment is causally derived from the simulated community dynamics.

\subsection{Supervised Fine-Tuning for Social Reasoning}

The first training stage involves Supervised Fine-Tuning (SFT) to teach the model the syntax and semantics of Social-CoT. We combine the heuristic-labeled Social-CoT data (from unlabeled UGC items) with human-annotated data into a unified corpus.
SFT plays a crucial role in enabling multi-modal grounding: it trains the model to align visual cues (e.g., lighting, editing pace) and textual metadata (titles, tags) with social interpretations. By learning to generate the reaction paths before predicting the label, the model internalizes a structured reasoning process, moving beyond black-box classification to interpretable social simulation.

\subsection{Process-Supervised Reinforcement Learning}

To further refine the quality of the Social-CoT generation, we employ Reinforcement Learning (RL) using Group Relative Policy Optimization (GRPO)~\cite{shao2024deepseekmathpushinglimitsmathematical}. While SFT teaches the model how to reason, RL aligns the reasoning process with authentic human social cognition. We design a composite reward signal comprising four distinct components:
\begin{equation}
    r = r_{\text{format}} + r_{\text{label}} + r_{\text{diversity}} + r_{\text{social}}.
\end{equation}

\paragraph{Format and Label Rewards.}
$r_{\text{format}}$ ensures the output adheres to the structured \texttt{<think>...</think>} format, while $r_{\text{label}}$ rewards the correctness of the final binary quality prediction against the ground truth.

\paragraph{Cognitive Diversity Constraint ($r_{\text{diversity}}$).}
A robust community simulation should reflect a spectrum of opinions rather than repeating a single viewpoint. To prevent mode collapse where the model generates repetitive comments, we introduce a diversity penalty:
\begin{equation}
    r_{\text{diversity}} = -\lambda_{\text{div}} \sum_{c \in \mathcal{C}} \ [f(c) - 1],
\end{equation}
where $\mathcal{C}$ is the set of generated reaction paths and $f(c)$ denotes the frequency of identical or near-identical sentiments, forcing the model to explore the full distribution of potential audience reactions.

\paragraph{Social Alignment Reward ($r_{\text{social}}$).}
To ensure the generated reasoning paths are not hallucinations but are grounded in genuine human emotional expression, we introduce the Social Alignment Reward, which measures the semantic similarity between the model's simulated personas and real, high-engagement user comments from a held-out set.
Let $\mathcal{G}=\{g_i\}$ be the set of generated reaction paths and $\mathcal{R}=\{r_j\}$ be the set of real user comments, we compute the cosine similarity between their embeddings:
\begin{equation}
    S_{ij} = e(g_i)^\top e(r_j), \quad \text{where } e(x) = \frac{f(x)}{\|f(x)\|_2}.
\end{equation}
We perform greedy matching to align each generated persona with the closest real-world counterpart. The final reward is the mean of these matched similarities:
\begin{equation}
    r_{\text{social}} = \frac{1}{|\mathcal{S}|} \sum_{s \in \mathcal{S}} s.
\end{equation}
This reward acts as a ``social grounding'' signal, encouraging the model to mimic the tone, nuance, and emotional granularity of actual community.

By combining all these rewards, the diversity and authenticity rewards ensure that simulated comments remain varied and semantically aligned with real user feedback, while the format and label rewards guarantee well-formed outputs and accurate final decisions. Together, these signals guide the model toward producing interpretable, community-grounded predictions for the CASTER task.

\section{Experiments}

In this section, we evaluate MEDEA on large-scale real-world UGC item assessment scenarios. We first introduce the experimental setups, including baselines and training data construction, followed by the main results on CASTER-Bench, and finally provide ablation studies to quantify the contribution of each system component.

\begin{table*}[t]
\centering
\small
\setlength{\tabcolsep}{5pt}
\begin{tabular}{lcccccccccc}
\toprule
\multirow{2}{*}{Method} 
& \multicolumn{3}{c}{High-Quality} 
& \multicolumn{3}{c}{Low-Quality} 
& \multicolumn{3}{c}{Macro Average} \\
\cmidrule(lr){2-4} \cmidrule(lr){5-7} \cmidrule(lr){8-10}
& Precision & Recall & F1 
& Precision & Recall & F1 
& Precision & Recall & F1 \\
\midrule
\multicolumn{10}{l}{\textit{\textbf{Traditional VQA Methods}}} \\
FastVQA~\cite{wu2022fastquality} &  0.347 &  0.440 &  0.388 &  0.761 &  0.682 &  0.719 &  0.554 &  0.561 &  0.554 \\
DOVER~\cite{wu2023exploring} &  0.308 &  0.377 &  0.339 &  0.739 &  0.676 &  0.706 &  0.524 &  0.526 &  0.523 \\
MaxVQA~\cite{wu2023towards} &  0.345 &  0.518 &  0.414 &  0.772 &  0.623 &  0.690 &  0.558 &  0.571 &  0.552 \\
Q-Align~\cite{wu2024q} & 0.382 & 0.404 & 0.392 & 0.766 & 0.749 & 0.758 & 0.574 & 0.577 & 0.575  \\
FineVQ~\cite{duan2025finevq} &  0.323 &  0.343 &  0.333 &  0.742 &  0.724 &  0.733 &  0.532 &  0.534 &  0.533 \\
VQA2~\cite{jia2025vqa2} & 0.358 & 0.454 & 0.400 & 0.766 & 0.688 & 0.725 & 0.562 & 0.571 & 0.562  \\
\midrule
\multicolumn{10}{l}{\textit{\textbf{Standard LMMs}}} \\
Qwen3-VL-Plus & 0.366 & 0.893 & 0.519 & 0.910 & 0.411 & 0.566 & 0.638 & 0.652 & 0.542 \\
GPT-5.2  & 0.347 & 0.933 & 0.506 & 0.929 & 0.332 & 0.489 & 0.638 & 0.633 & 0.498 \\
Claude-4.5-opus & 0.309 & \textbf{0.995} & 0.472 & \textbf{0.988} & 0.148 & 0.257 & 0.648 & 0.571 & 0.364  \\
\midrule
\multicolumn{10}{l}{\textit{\textbf{Reasoning-Enhanced LMMs (Long-CoT)}}} \\
Qwen3-VL-8B-Think (backbone) & 0.265 & 0.115 & 0.160 & 0.721 & \textbf{0.892} & 0.797 & 0.493 & 0.504 & 0.479 \\
Qwen3-VL-Plus (reasoning)  & 0.316 & 0.905 & 0.468 & 0.872 & 0.247 & 0.385 & 0.594 & 0.576 & 0.427 \\
GPT-5.2 (reasoning)  & 0.401 & 0.903 & 0.555 & 0.928 & 0.483 & 0.635 & 0.665 & 0.693 & 0.595 \\
Gemini-3.0-Pro (reasoning)  & 0.313 & 0.978 & 0.474 & 0.954 & 0.176 & 0.297 & 0.634 & 0.577 & 0.385 \\
Claude-4.5-opus (reasoning)  & 0.364 & 0.964 & 0.528 & 0.962 & 0.353 & 0.517 & 0.663 & 0.658 & 0.522  \\
\midrule
\multicolumn{10}{l}{\textit{\textbf{Flagship Models with Social-CoT Simulation}}} \\
Gemini-2.5-Flash (social-CoT) & 0.353 & 0.629 & 0.452 & 0.779 & 0.615 & 0.687 & 0.566 & 0.622 & 0.570 \\
Qwen3-VL-Plus (social-CoT)  & 0.380 & 0.766 & 0.508 & 0.853 & 0.521 & 0.647 & 0.617 & 0.644 & 0.578 \\
GPT-5.2 (social-CoT) & 0.442 & 0.304 & 0.360 & 0.762 & 0.853 & 0.805 & 0.602 & 0.578 & 0.582 \\
Claude-4.5-opus (social-CoT)  & 0.371 & 0.810 & 0.510 & 0.867 & 0.474 & 0.613 & 0.619 & 0.642 & 0.561  \\
\midrule
\textbf{MEDEA (Ours)} & \textbf{0.603} & 0.705 & \textbf{0.650} & 0.850 & 0.845 & \textbf{0.847} & \textbf{0.727} & \textbf{0.775} & \textbf{0.749} \\
\bottomrule
\end{tabular}
\caption{Main results on CASTER-Bench. We compare MEDEA against four categories of baselines: Traditional VQA, Standard LMMs, Reasoning-Enhanced LMMs (Long-CoT), and Social-CoT simulated models. We report precision, recall, and F1-score for the High-Quality and Low-Quality classes, as well as macro-averaged metrics. Since the CASTER task focuses on identifying truly high-quality content from high-exposure UGC, performance on the High-Quality class is particularly critical. Token overhead and reasoning cost are presented in Appendix~\ref{app:inference_cost}.}
\label{tab:main_results}
\end{table*}

\subsection{Experimental Setups}

\paragraph{Baselines.}
To comprehensively assess the performance of MEDEA, we compare it against a diverse set of baselines categorized into four groups:

\begin{enumerate}[nosep]
    \item \textbf{Traditional Video Quality Assessment (VQA) Methods:} This group includes representative regression-based models that focus on aesthetic and technical quality, including FastVQA~\cite{wu2022fastquality}, DOVER~\cite{wu2023exploring}, MaxVQA~\cite{wu2023towards}, Q-Align~\cite{wu2024q}, FineVQ~\cite{duan2025finevq}, and VQA2~\cite{jia2025vqa2}.
    \item \textbf{Standard Large Multimodal Models (LMMs):} We evaluate general-purpose flagship models, including Qwen3-VL-Plus, GPT-5.2 and Claude-4.5-opus. These LMMs are the only flagship candidates capable of explicitly disabling the reasoning process, allowing us to establish a pure baseline for standard multimodal capabilities without intrinsic CoT interference.
    \item \textbf{Reasoning-Enhanced LMMs (Long-CoT):} To benchmark against state-of-the-art intrinsic reasoning capabilities, we include models utilizing CoT or long-context reasoning. This category includes Qwen3-VL-8B-Think (the backbone of MEDEA), Qwen3-VL-Plus (reasoning), GPT-5.2 (reasoning), Gemini-3.0-Pro (reasoning), and Claude-4.5-opus (reasoning). For these models, we explicitly configured the reasoning effort to ``high'' to fully activate their extended thinking capabilities and maximize the depth of logical deduction.
    \item \textbf{Flagship Models with Social-CoT Simulation:} To isolate the effectiveness of our proposed mechanism, we prompt non-reasoning models (Gemini-2.5-Flash, Qwen3-VL-Plus, and GPT-5.2) with the Social-CoT prompts used in MEDEA, forcing them to simulate social perspective-taking without fine-tuning.
\end{enumerate}

For the Traditional VQA methods, which output continuous quality scores, we perform an exhaustive threshold sweep to map scores to binary classifications and report the best performance on CASTER-Bench, ensuring they are evaluated at their optimal operating points. Detailed results of these baselines across various thresholds are provided in Appendix~\ref{app:detailed_results}. All LMM-based baselines perform zero-shot prediction. Flagship Models with Social-CoT Simulation utilize the exact inference prompt as MEDEA to ensure a fair comparison of the reasoning framework itself. All reported results are averaged over five independent runs.

\paragraph{Training Data.}
The full data construction pipeline is described in Section~\ref{sec:data_construction}. Here we summarize key components. For unlabelled UGC items, we query Gemini-2.5-Flash to generate reasoning traces and pseudo-labels. The model receives multimodal and metadata-rich inputs, including Cover image, 7 key frames sampled from the video, Title, Tags, ASR transcript, Primary category label, Secondary category label, Video duration, Resolution, Orientation (vertical / non-vertical) and Top 50 most-liked comments from which 15–20 content-relevant comments are selected. This process yields 54k Gemini-labeled CoT samples. For the 3k human-annotated UGC items, we additionally supply the ground-truth quality label when prompting Gemini, enabling it to generate supervision traces aligned with human judgment. Prompt templates used for CoT generation are provided in Appendix~\ref{app:prompts}. During SFT, we train MEDEA on the combined Gemini-labeled and human-annotated corpus. During RL, we only use the human-curated samples, ensuring that the reinforcement signal is anchored to expert-quality annotations. Additional training configurations and hyperparameters are also included in Appendix~\ref{sec:appendix_exp_args}.

\subsection{Main Results}

Table~\ref{tab:main_results} presents the main results on CASTER-Bench. A defining property of this benchmark is its imbalanced label distribution: High-Quality UGC constitutes only a small fraction of the data. Consequently, performance on the High-Quality class is the most critical metric, as it reflects a model's ability to recognize intrinsic excellence rather than merely filtering out obvious failures.

MEDEA demonstrates superior performance, significantly outperforming all baselines across every category. It achieves an F1 score of 0.650 on the High-Quality class, surpassing the strongest baseline by a large margin. Crucially, MEDEA strikes an optimal balance between precision (0.603) and recall (0.705). This indicates strong selectivity—a capability essential for practical recommendation systems where false positives degrade user trust.

Analyzing the baseline categories reveals distinct failure modes:

\paragraph{Generosity Bias in LMMs.} A striking phenomenon is observed in both Standard LMMs and Reasoning-Enhanced LMMs. Flagship models like GPT-5.2 and Claude-4.5-Opus achieve near-perfect Recall ($>90\%$) on the High-Quality class but suffer from extremely low Precision ($\sim30\%$). This suggests that while these models can identify positive attributes in almost any video via long-context reasoning, they exhibit a "Generosity Bias". They tend to over-rationalize merit in average content, lacking the critical social discernment to distinguish "acceptable" content from "community-resonant" masterpieces.

\paragraph{Signal-Dominance in Traditional VQA.} Traditional methods (e.g., FastVQA, VQA2) are heavily biased towards Low-Quality class. Their High-Quality F1 scores remain consistently poor (ranging from 0.33 to 0.41), confirming that aesthetic fidelity alone is insufficient for capturing the semantic and social dimensions of community resonance.

\paragraph{Effectiveness of Social Alignment.} While prompting flagship models with Social-CoT (the fourth category) improves performance over standard zero-shot inference, they still lag behind MEDEA. For instance, Qwen3-VL-Plus with Social-CoT achieves an F1 of 0.508 compared to MEDEA's 0.650. This validates that the reasoning pattern alone is not enough; the model requires the specific alignment with expert-curated social judgments provided by MEDEA's training pipeline to internalize the true "community standard".

Finally, MEDEA achieves the highest Macro-F1 score (0.749), reflecting robust performance across the entire quality spectrum. Its ability to maintain high recall without succumbing to the positivity bias of general-purpose reasoning models validates the effectiveness of the proposed framework.

\subsection{Ablation Experiments}
\label{sec:ablation}

\begin{table*}[t]
\centering
\small
\setlength{\tabcolsep}{5pt}
\begin{tabular}{lcccccccccc}
\toprule
\multirow{2}{*}{Method} 
& \multicolumn{3}{c}{High-Quality} 
& \multicolumn{3}{c}{Low-Quality} 
& \multicolumn{3}{c}{Macro Average} \\
\cmidrule(lr){2-4} \cmidrule(lr){5-7} \cmidrule(lr){8-10}
& Precision & Recall & F1 
& Precision & Recall & F1 
& Precision & Recall & F1 \\
\midrule
SFT-pseudo-label & 0.379 & 0.683 & 0.487 & 0.802 & 0.599 & 0.686 & 0.591 & 0.641 & 0.587 \\
SFT-human-label & 0.341 & 0.407 & 0.371 & 0.755 & 0.670 & 0.710 & 0.548 & 0.539 & 0.541 \\
SFT-w/o-social-CoT & 0.377 & 0.787 & 0.510 & 0.833 & 0.517 & 0.638 & 0.605 & 0.652 & 0.574 \\
\textbf{SFT-pseudo+human} & 0.398 & 0.750 & 0.520 & 0.829 & 0.557 & 0.666 & 0.614 & 0.654 & 0.593 \\
\midrule
RL-pseudo+human & 0.578 & 0.500 & 0.536 & 0.790 & 0.916 & 0.848 & 0.684 & 0.708 & 0.692 \\
RL-w/o-social-reward & 0.575 & 0.657 & 0.613 & 0.836 & 0.837 & 0.836 & 0.706 & 0.747 & 0.725 \\
RL-w/o-social-CoT  & 0.504 & 0.362 & 0.421 & 0.770 & 0.880 & 0.821 & 0.637 & 0.621 & 0.621 \\
\textbf{MEDEA(RL-human-label)} & 0.603 & 0.705 & 0.650 & 0.850 & 0.845 & 0.847 & 0.727 & 0.775 & 0.749 \\
\bottomrule
\end{tabular}
\caption{Ablation studies on CASTER-Bench. Each component of MEDEA contributes to overall performance.}
\label{tab:ablation_results}
\end{table*}

To isolate the contribution of each component in MEDEA, we perform a series of ablations. Specifically, we analyze the impact of the Social-CoT and the Social Alignment Reward (denoted as $r_{social}$).

\paragraph{Necessity of Social Reasoning Paths.}
Removing the Social-CoT (``RL-w/o-social-CoT'') leads to a substantial performance drop, with the High-Quality F1 score decreasing from 0.650 to 0.421. This sharp decline confirms that pixel-level perception alone is insufficient for assessing community resonance. The Social-CoT acts as a necessary cognitive bridge, allowing the model to perform multimodal perspective-taking to infer how content features translate into user engagement.

\paragraph{Impact of Social Alignment and Qualitative Analysis.} Excluding the Social Alignment Reward leads to \textit{Social Mode Collapse}, where reasoning degenerates into repetitive, generic templates (e.g., "So beautiful"). Qualitative inspection in Appendix~\ref{sec:appendix_qualitative} confirms this distinction: while MEDEA empathetically interprets wind-swept keyframes in an Iceland vlog as "raw natural power", the ablated model produces only hollow praise. This underscores that social alignment is critical for grounding the model in authentic, emotionally nuanced community expression.



\section{Conclusions}
This work establishes a new paradigm for UGC assessment, shifting focus from aesthetic fidelity to social-cognitive resonance. By introducing the Social-CoT mechanism, we demonstrate that effective quality assessment requires not just signal analysis, but the capacity for multimodal perspective-taking. Our framework, MEDEA, validates that simulating a "community of critics" via Social Alignment Reward effectively captures the nuance of human engagement. Beyond specific performance gains on CASTER-Bench, this research paves the way for equipping LMMs with Theory of Mind capabilities, bridging the gap between computational metrics and genuine social understanding.

\section*{Limitations}
While MEDEA demonstrates strong performance on community-aware UGC assessment, several limitations remain. First, although the Social-CoT mechanism incurs additional computational cost compared to direct prediction (as detailed in Appendix~\ref{app:inference_cost}), this overhead is slightly higher than that of some reasoning-enhanced LMMs, but since MEDEA has a much smaller parameter size, the overall cost and inference time remain controllable. Second, the current social alignment is optimized for specific platform dynamics; consequently, its generalizability to other social ecosystems with distinct cultural norms or community behaviors remains to be verified. Third, our binary framing oversimplifies the continuous spectrum of community resonance. Finally, while our current implementation leverages rich multimodal metadata for social grounding, the MEDEA framework is theoretically extensible to single-modality or sparse-signal scenarios, which we leave for future exploration.


\bibliographystyle{plain}
\bibliography{refs}

\newpage

\appendix

\section{Token Overhead and Reasoning Cost}
\label{app:inference_cost}

Table~\ref{tab:token_cost} details the computational overhead associated with the reasoning process. Integrating the Social-CoT module significantly increases the generation volume: MEDEA generates an average of 1,256 tokens per UGC item, compared to just 5.6 tokens for the direct-answer variant (MEDEA w/o Social-CoT).

\paragraph{Inference Efficiency.} We evaluate efficiency on local 4$\times$H800 GPUs using vLLM with 8 concurrent workers. The generation of dense social reasoning reduces inference throughput from 2.55 to 0.79 videos/sec. However, this increased latency is a necessary trade-off for precision. As shown in Table~\ref{tab:main_results}, this computational investment yields a High-Quality F1 score of 0.650, outperforming the fastest traditional methods (F1 $\approx$ 0.33--0.41) which fail to capture semantic resonance.

\paragraph{Comparison with Reasoning Baselines.} Analyzing the relationship between token consumption and performance reveals that simply increasing reasoning length does not guarantee better judgment:

\begin{itemize}
\item \textbf{Inefficient Deep Reasoning:} High token consumption does not automatically translate to high accuracy. For instance, Qwen3-VL-Plus (reasoning) generates nearly 1,000 tokens per video (917.5) but only achieves a High-Quality F1 of 0.468. Despite a reasoning depth comparable to ours, it lacks the specific social alignment, resulting in verbose but ultimately misaligned judgments that succumb to the generosity bias.

\item \textbf{Shallow Reasoning Limits:} Conversely, models with lower reasoning overheads, such as GPT-5.2 (reasoning) and Gemini-3.0-Pro (reasoning), consume significantly fewer tokens (96.5 and 160.0, respectively). However, this efficiency caps their performance (High-Quality F1 of 0.555 and 0.474), suggesting that the complex social dynamics of UGC cannot be adequately captured through brief, surface-level chain-of-thought processes.

\item \textbf{Simulation vs. Alignment:} Flagship models prompted with Social-CoT (e.g., Claude-4.5-Opus at 712.4 tokens) sit in the middle ground, utilizing moderate token budgets to simulate social critique. Yet, they still fall short of MEDEA (F1 0.510 vs. 0.650). This indicates that MEDEA's higher token count (1,256) is not merely verbose, but represents a necessary depth of analysis derived from training on expert data—internalizing a standard that prompt engineering alone cannot fully replicate.
\end{itemize}

In summary, MEDEA leverages a higher token budget to construct a critical social context that other models either gloss over (shallow reasoners) or misinterpret through excessive positivity (deep reasoners).

\begin{table}[t]
\centering
\small
\setlength{\tabcolsep}{6pt}
\begin{tabular}{lcc}
\toprule
Method & Videos/sec & Tokens\\
\midrule
\multicolumn{3}{l}{\textit{\textbf{Ours (Local Inference)}}} \\
MEDEA-w/o-social-CoT & 2.55 & 5.6 \\
MEDEA & 0.79 & 1,256.0 \\
\midrule
\multicolumn{3}{l}{\textit{\textbf{Reasoning-Enhanced LMMs (API-based)}}} \\
GPT-5.2 (reasoning) & - & 96.5 \\
Gemini-3.0-Pro (reasoning) & - & 160.0  \\
Claude-4.5-Opus (reasoning) & - & 563.4 \\
Qwen3-VL-Plus (reasoning) & - & 917.5 \\
\midrule
\multicolumn{3}{l}{\textit{\textbf{Social-CoT Simulation (API-based)}}} \\
Qwen3-VL-Plus (social-CoT) & - & 424.9  \\
GPT-5.2 (social-CoT) & - & 489.2 \\
Claude-4.5-Opus (social-CoT) & - & 712.4 \\
\bottomrule
\end{tabular}
\caption{Average tokens per UGC item and inference efficiency. MEDEA’s higher token count reflects the generation of dense social context, which is critical for High-Quality identification. Baselines are API-based; speed/hardware not reported.}
\label{tab:token_cost}
\end{table}

\section{Hyperparameters used in training and inference of MEDEA}
\label{sec:appendix_exp_args}
Hyperparameters used in training and inference of MEDEA are presented in Table~\ref{Table:hyper}.

\begin{table}
\centering
\small
\begin{tabular}{clc} 
\toprule
{Stage} & {Hyperparameter} & {Value} \\

\midrule
\multirow{4}{*}{SFT}
& batch size & 256 \\
& learning rate & 5e-6 \\
& learning rate schedule & cosine \\
& learning rate decay ratio & 0.2 \\

\midrule
\multirow{11}{*}{RL}
& batch size & 64 \\
& learning rate & 1e-6 \\
& learning rate schedule & cosine \\
& learning rate decay ratio & 0.1 \\
& PPO clip ratio low & 0.2 \\
& PPO clip ratio high & 0.2 \\
& kl coefficient & 0.001 \\
& entropy coefficient & 0.001 \\
& rollout number & 8 \\
& rollout top-p & 1.0 \\
& rollout temperature & 0.6 \\
& rollout repetition penalty & 1.0 \\

\midrule
\multirow{4}{*}{Inference}
& top-k & 50 \\
& top-p & 0.7 \\
& temperature & 0.6 \\
& repetition penalty & 1.0 \\

\bottomrule
\end{tabular}
\caption{Hyperparameters used in training and inference of MEDEA.}
\label{Table:hyper}
\end{table}

\section{Modality Ablation: Text-Only vs. Vision-Only}
\label{sec:appendix_modality_ablation}

To better understand the contribution of different modalities, we conduct a systematic ablation study comparing three settings: Text-Only, Vision-Only, and the full multimodal MEDEA. The Text-Only setting uses title, tags, ASR transcripts, and metadata, without any visual input. The Vision-Only setting uses the cover image and sampled key frames, without textual inputs. The full MEDEA model leverages both modalities. Here are some findings: 

\begin{table*}[t]
\centering
\small
\begin{tabular}{lccccccccc}
\toprule
\textbf{Setting} 
& \textbf{HQ-P} & \textbf{HQ-R} & \textbf{HQ-F1} 
& \textbf{LQ-P} & \textbf{LQ-R} & \textbf{LQ-F1} 
& \textbf{Avg-P} & \textbf{Avg-R} & \textbf{Avg-F1} \\
\midrule
Text-Only   & 0.511 & 0.703 & 0.592 & 0.867 & 0.742 & 0.803 & 0.689 & 0.723 & 0.698 \\
Vision-Only & 0.571 & 0.487 & 0.526 & 0.814 & 0.860 & 0.836 & 0.693 & 0.673 & 0.681 \\
MEDEA       & 0.603 & 0.705 & 0.650 
            & 0.850 & 0.845 & 0.847 
            & 0.727 & 0.775 & 0.749 \\
\bottomrule
\end{tabular}
\caption{Modality ablation results on CASTER-Bench.}
\label{tab:modality_ablation}
\end{table*}

\paragraph{Neither modality alone is sufficient.}
Text-Only achieves a Macro-F1 of 0.698, and Vision-Only achieves 0.681, both significantly lower than MEDEA (0.749). This indicates that CASTER cannot be effectively solved using a single modality.

\paragraph{Complementary strengths of text and vision.}
Text-Only achieves higher HQ-Recall (0.703) but lower Precision (0.511), suggesting that textual signals are effective for identifying potential high-quality candidates but are prone to false positives (e.g., clickbait or misleading titles). 
In contrast, Vision-Only achieves higher Precision (0.571) but lower Recall (0.487), indicating that visual signals provide more reliable confirmation of quality but may miss cases where engagement is driven by narrative or semantic content.
MEDEA effectively combines these complementary strengths.

\paragraph{Both modalities are indispensable.}
Removing visual input (Text-Only vs. MEDEA) leads to a drop of 5.1 points in Macro-F1, while removing textual input (Vision-Only vs. MEDEA) results in a larger drop of 6.8 points.
This demonstrates that both modalities play critical and non-redundant roles in modeling community resonance.

\section{Faithfulness and Diversity of Generated Reasoning}
\label{sec:appendix_faithfulness}

Hallucinated or weakly grounded reasoning is a known risk in multimodal reasoning models. In MEDEA, this issue is mitigated through multimodal grounding (conditioning on frames, ASR, and metadata) and a \emph{Think-then-Aggregate} structure that enforces internal consistency across reasoning paths. To systematically evaluate reasoning quality, we conduct an external blind assessment using Gemini as an independent judge. We randomly sample outputs from two variants: MEDEA w/o $r_{\text{social}}$ and the full MEDEA model. For each sample, the judge is provided with the original multimodal inputs (video summary, frames, ASR, metadata) together with the generated reasoning paths and simulated comments, and rates them on a \textbf{5-point scale} (1 = very poor, 5 = excellent) along two dimensions: \textbf{Faithfulness} (grounding in observable video evidence) and \textbf{Diversity} (variation and non-redundancy across perspectives). The evaluation is conducted blindly without revealing model identity.

\begin{table}[t]
\centering
\small
\begin{tabular}{lcc}
\toprule
\textbf{Setting} & \textbf{Faithfulness (↑)} & \textbf{Diversity (↑)} \\
\midrule
MEDEA w/o $r_{\text{social}}$ & 2.471 & 1.058 \\
MEDEA & \textbf{4.211} & \textbf{2.743} \\
\bottomrule
\end{tabular}
\caption{Evaluation of reasoning faithfulness and diversity (5-point scale).}
\label{tab:faithfulness_diversity}
\end{table}

The results show that incorporating the Social Alignment Reward substantially improves both faithfulness and diversity. The full MEDEA model achieves stronger grounding in video content and produces more varied and less redundant perspectives. Qualitative inspection further indicates that removing the reward leads to generic and repetitive reasoning patterns with weaker alignment to specific narrative elements, while the full model more frequently references concrete visual and ASR cues. These findings suggest that the Social Alignment Reward enhances structured, grounded, and socially coherent reasoning rather than merely promoting stylistic variation.

\section{Qualitative Analysis of Social Reasoning Paths}
\label{sec:appendix_qualitative}

To qualitatively illustrate how the Social-CoT mechanism instantiates diverse viewer personas to achieve social reasoning, we present representative examples of reasoning paths under three settings:

\begin{enumerate}
    \item \textbf{Oracle Social Context} (Figure~\ref{fig:cot_gemini}): Social-CoT generated by a strong proprietary model (Gemini) conditioned on \emph{real, high-engagement user comments}. This serves as the ``upper bound'' or gold standard for community-aligned reasoning.
    \item \textbf{Social-CoT with Alignment} (Figure~\ref{fig:cot_medea}): Reasoning paths generated by MEDEA using the full Social Alignment Reward ($r_{social}$). This demonstrates the model's capability for \textit{Empathetic Simulation}.
    \item \textbf{Social-CoT without Alignment} (Figure~\ref{fig:cot_medea_no_real}): Reasoning paths generated by MEDEA without the social alignment constraint. This illustrates the phenomenon of ``Social Mode Collapse'', where reasoning becomes repetitive and robotic.
\end{enumerate}

\begin{figure*}[t]
    \centering
    \begin{subfigure}{0.24\textwidth}
        \centering
        \includegraphics[width=\linewidth]{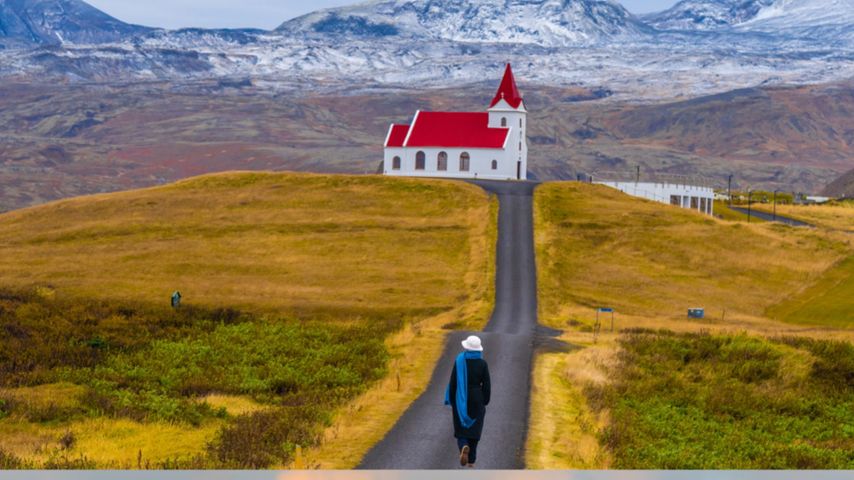}
        \caption{Cover}
        \label{fig:frame_8}
    \end{subfigure}
    \hfill
    \begin{subfigure}{0.24\textwidth}
        \centering
        \includegraphics[width=\linewidth]{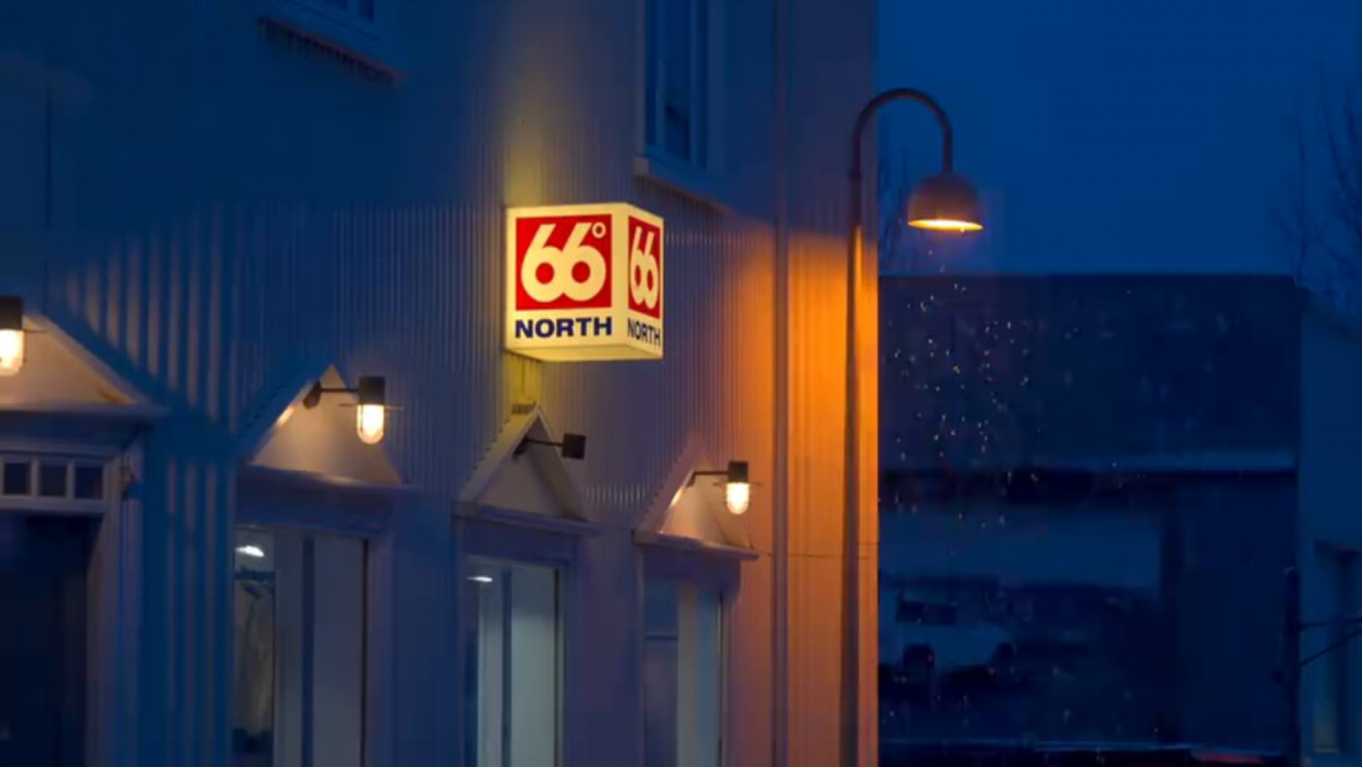}
        \caption{Key Frame 1}
        \label{fig:frame_1}
    \end{subfigure}
    \hfill
    \begin{subfigure}{0.24\textwidth}
        \centering
        \includegraphics[width=\linewidth]{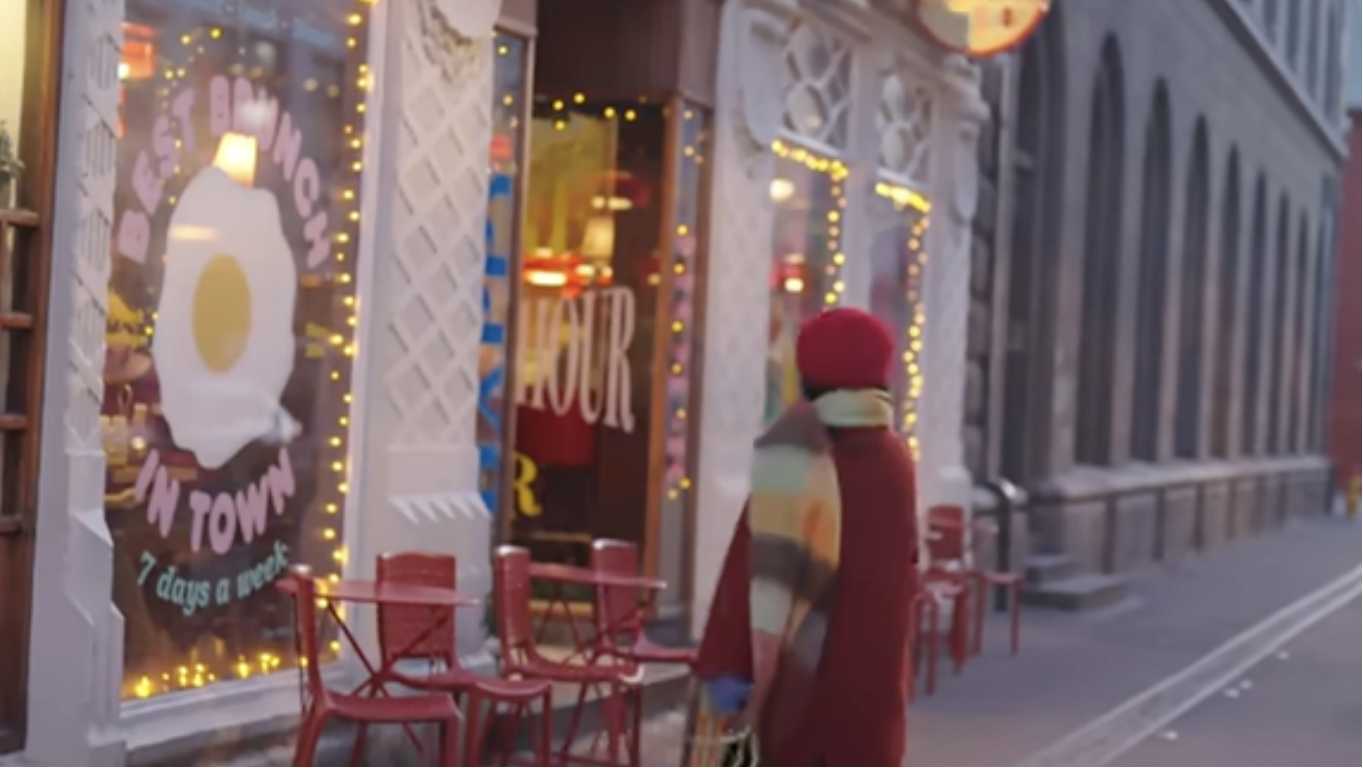}
        \caption{Key Frame 2}
        \label{fig:frame_2}
    \end{subfigure}
    \hfill
    \begin{subfigure}{0.24\textwidth}
        \centering
        \includegraphics[width=\linewidth]{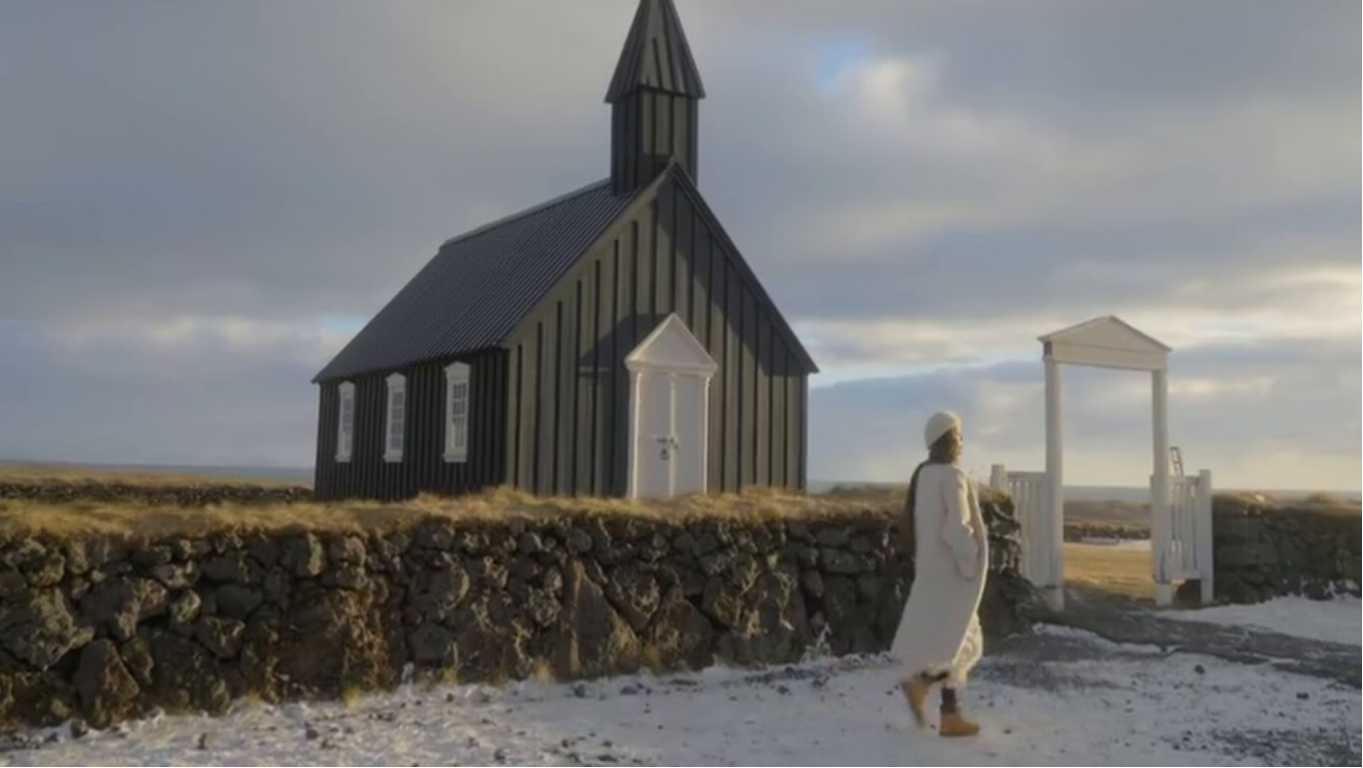}
        \caption{Key Frame 3}
        \label{fig:frame_3}
    \end{subfigure}
    \hfill
    \begin{subfigure}{0.24\textwidth}
        \centering
        \includegraphics[width=\linewidth]{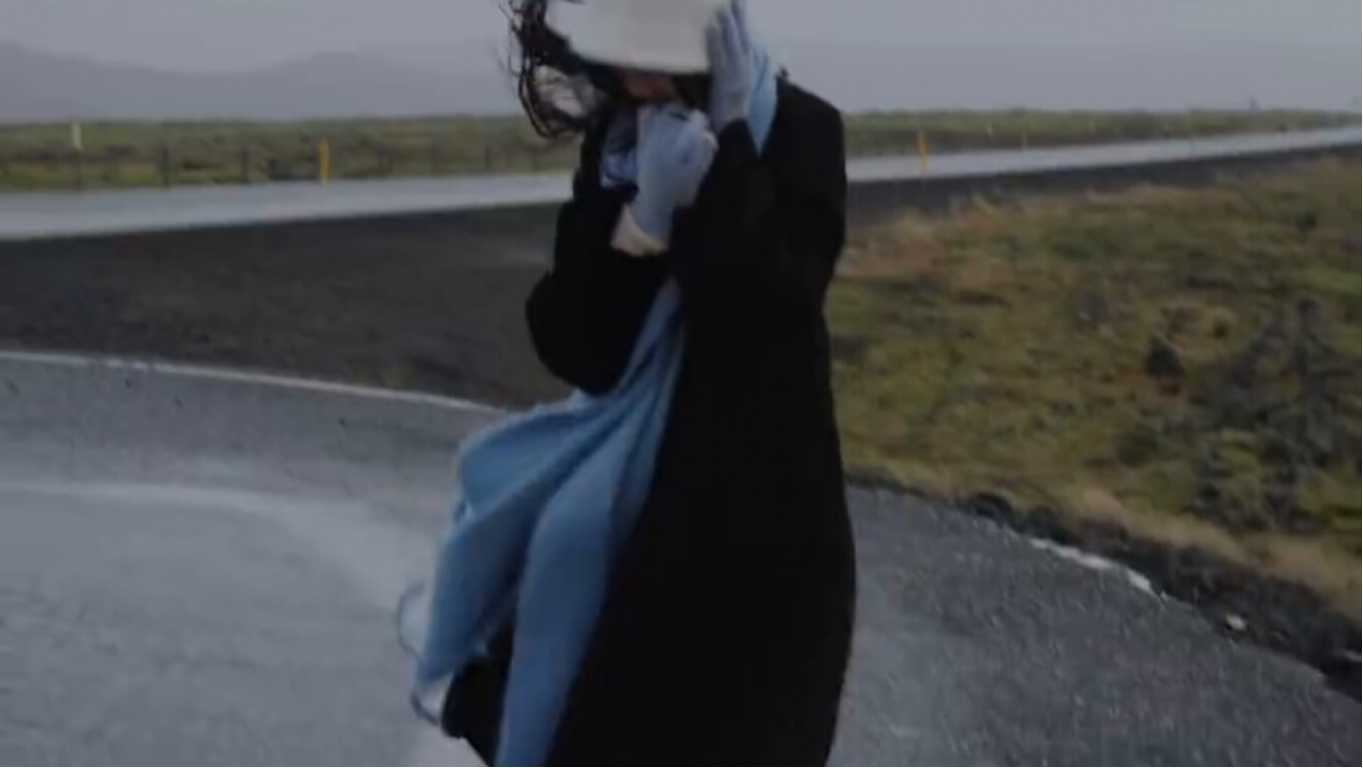}
        \caption{Key Frame 4}
        \label{fig:frame_4}
    \end{subfigure}
    \hfill
    \begin{subfigure}{0.24\textwidth}
        \centering
        \includegraphics[width=\linewidth]{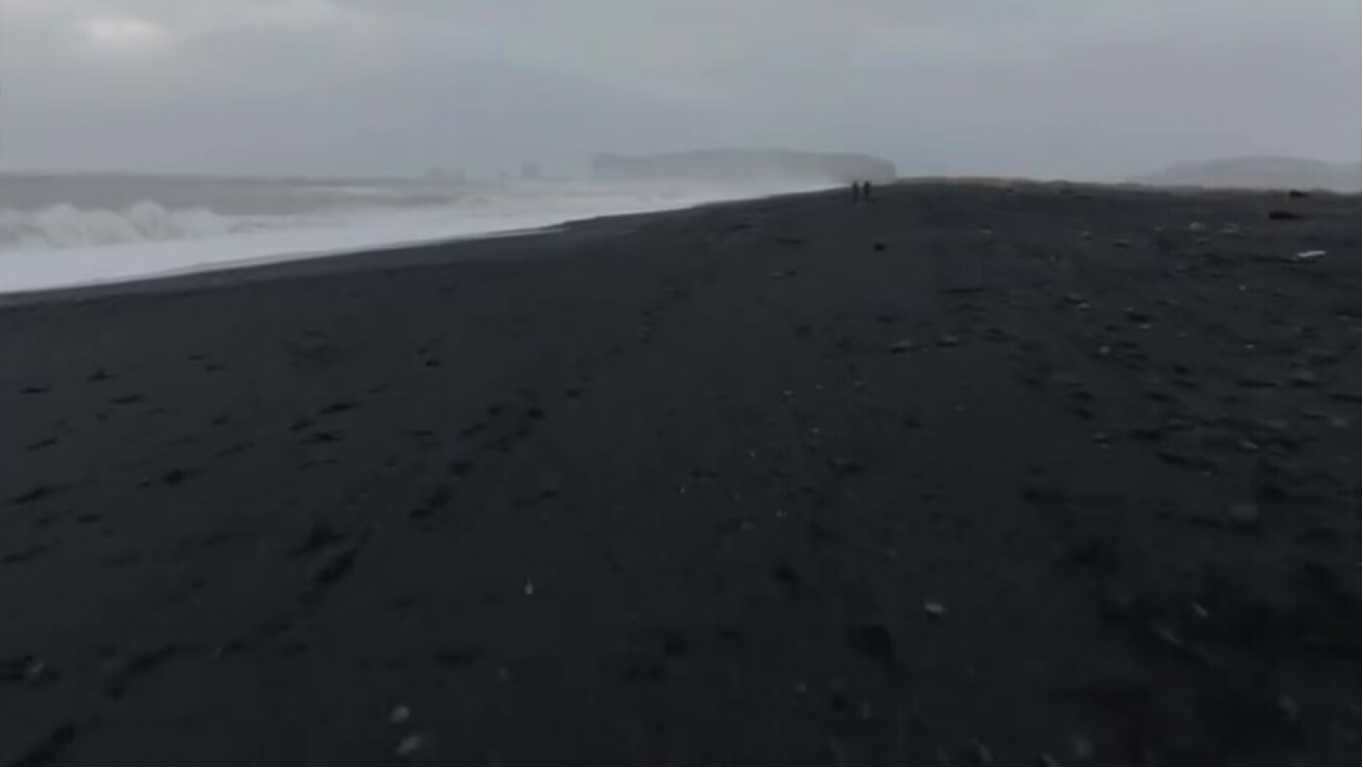}
        \caption{Key Frame 5}
        \label{fig:frame_5}
    \end{subfigure}
    \hfill
    \begin{subfigure}{0.24\textwidth}
        \centering
        \includegraphics[width=\linewidth]{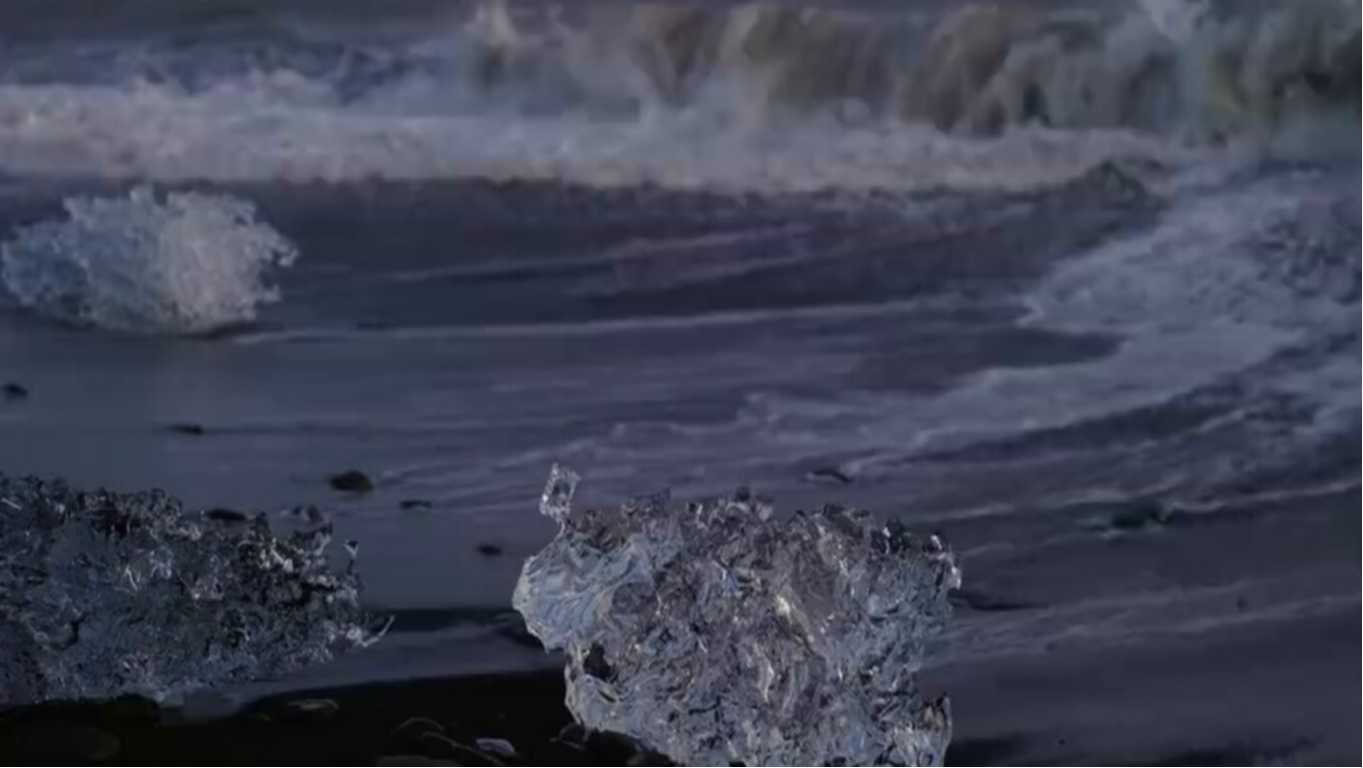}
        \caption{Key Frame 6}
        \label{fig:frame_6}
    \end{subfigure}
    \hfill
    \begin{subfigure}{0.24\textwidth}
        \centering
        \includegraphics[width=\linewidth]{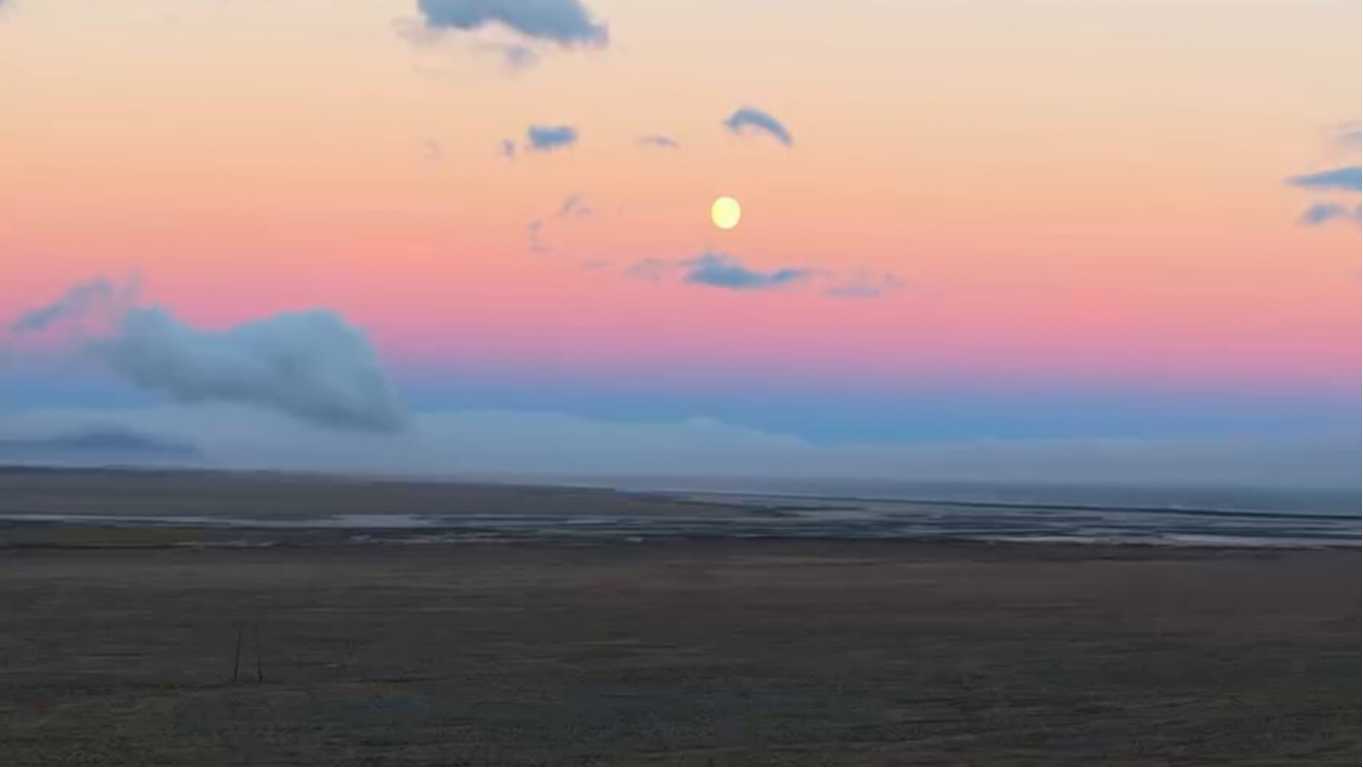}
        \caption{Key Frame 7}
        \label{fig:frame_7}
    \end{subfigure}
    
    \caption{Cover and 7 uniformly sampled key frames of the example.}
    \label{fig:example_images}
\end{figure*}

We additionally provide the UGC item cover image together with seven uniformly sampled key frames in Figure~\ref{fig:example_images}, which serve as the visual context available to the model during perspective-taking. These frames capture representative scenes, visual quality, and narrative progression, enabling readers to assess how well the generated Social-CoT aligns with the visual narrative.

\paragraph{Analysis of Oracle Social Context.}
The first setting (Figure~\ref{fig:cot_gemini}) serves as a reference for authentic social cognition. By accessing real community feedback, the reasoning path exhibits rich, fine-grained emotional granularity, connecting specific visual metaphors (e.g., ``The wilderness is a determination'') to deep philosophical reflections found in the comment section.

\paragraph{Analysis of Social Alignment.}
The comparison between MEDEA with and without Social Alignment highlights the emergence of social intelligence.

As illustrated in the case study of an Iceland trip vlog (Figure~\ref{fig:cot_medea}), MEDEA demonstrates the ability to simulate empathy. Instead of merely listing technical attributes like resolution or lighting, the model instantiates diverse viewer personas to evaluate the content's visual narrative. For instance, by analyzing key frames that depict people walking against strong gusts, the model interprets this not just as motion, but as a manifestation of Iceland's raw natural power. It consequently simulates a viewer's visceral reaction: "The wind in Iceland looks intense, really shocking". This indicates that MEDEA has internalized the nuanced, multi-faceted "voice" of the community.

In stark contrast, Figure~\ref{fig:cot_medea_no_real} (Without Alignment) demonstrates Social Mode Collapse. While the model correctly identifies the content as ``beautiful'', the reasoning path degenerates into repetitive templates (e.g. repeating ``So beautiful... I really want to go'' multiple times). This confirms that without the Social Alignment Reward, the model fails to capture the diverse ``voice'' of the community, resulting in a hollow simulation lacking empathetic depth.

Overall, these examples demonstrate that Social-CoT can effectively substitute real user feedback in driving engagement-aware reasoning, and that the Social Alignment Reward plays a crucial role in improving the authenticity, coherence, and interpretability of the generated reasoning process.

\section{Prompts used in MEDEA}
\label{app:prompts}

We present the complete prompt used to instruct the teacher model to perform comment selection, stance classification, and reasoning-based aggregation for UGC items. The prompt is designed to simulate how users infer the creative quality of a UGC item from its visual and textual content, and how such inferences are reflected in the comment section.

The task formulation explicitly constrains the model to rely only on observable video attributes, including the cover image, key frames, metadata, and automatically transcribed text, while excluding any auditory or external signals. To ensure interpretability and reproducibility, the prompt enforces strict rules on comment selection, independent coverage of each comment, and a final statistically grounded stance decision based on a Skellam-normalized difference score. The prompt used to generate reasoning content is presented in Figure~\ref{fig:gemini}.

We design a structured prompt to guide MEDEA in simulating comment-section reactions on UGC items. The prompt integrates both visual inputs (cover image and key frames) and textual metadata (title, tags, ASR, category, and video attributes), encouraging the model to reason about the perceived creation quality of a UGC item. Instead of directly predicting an overall label, the model is required to first generate a diverse set of stance-aware comments.
The final judgment is derived through a quantitative aggregation process based on a Skellam $z$-score, which measures the normalized difference between supportive and opposing comments. This design enforces internal consistency, reduces shortcut learning, and aligns the prediction with interpretable intermediate reasoning. The prompt used to train MEDEA is presented in Figure~\ref{fig:medea}.

\section{Statistical Significance Testing}
To more comprehensively evaluate the performance of our MEDEA method, we incorporated p-values alongside conventional metrics in Table~\ref{tab:pvalue}. The consistent statistical significance observed across all experimental results, as clearly demonstrated in the accompanying table, strongly attests to the robustness of our approach. These findings not only provide compelling evidence that our method substantially outperforms the baseline but also highlight its reliability and generalizability under varied conditions.

\begin{table}[t]
\centering
\small
\setlength{\tabcolsep}{5pt}
\begin{tabular}{lccc}
\toprule
\multirow{2}{*}{Method} 
& \multicolumn{3}{c}{Macro Average} \\
\cmidrule(lr){2-4}
& Precision & Recall & F1 \\
\midrule
MEDEA & 5.0e-04 & 3.4e-02 & <1.0e-10 \\

\bottomrule
\end{tabular}
\caption{P-values comparing MEDEA with the best baseline (GPT-5.2 reasoning) using paired bootstrap tests.}
\label{tab:pvalue}
\end{table}

\section{Distinguishing Intrinsic Quality from Popularity}
\label{app:intrinsic_vs_popularity}
In this study, the core objective of the CASTER task is to assess the intrinsic value of UGC items, rather than merely predicting their current popularity, which is influenced by various external factors. The expert-annotated dataset we employ essentially serves as a "refinement" and "correction" of the noisy real-world community signals. Authentic user interaction data is saturated with noise, such as click-farming bots, irrational herd behavior, and biases inherent in the platform’s recommendation algorithms. Therefore, expert annotations provide a well-considered and idealized signal based on the intrinsic value of the content itself.

To illustrate this point more tangibly, we present some representative cases observed in the dataset in Table~\ref{fig:cases}, which demonstrate the fundamental distinction between learning from expert judgments and blindly fitting popularity metrics. Certain UGC items with high actual view counts or interaction metrics are labeled as low-quality by experts. Such content often relies on sensationalist titles, vulgar visual elements, or misleading information, with high traffic stemming more from emotional provocation or short-term platform recommendation strategies than intrinsic value.

By training models to fit this "refined" expert signal, the CASTER task aims to advance the modeling and recognition of content quality itself.

\begin{figure*}[t]
\centering
\begin{tcolorbox}[
  width=\textwidth,
  colback=white,
  colframe=black,
  title=Cases of High-Popularity but Low-Expert-Rated Content,
  listing options={
    basicstyle=\small\ttfamily,
    breaklines=true
  }
]
\small
Example1\\

\textbf{Title}: \\
The Unboxing of WLOP's Art Collection "Ghostblade 4" Is Here!! \textbf{There's a Giveaway}!

\textbf{Tag}: \\
Unboxing, Sharing, Ghostblade, Artbook, WLOP, Ghostblade 4

\textbf{ASR}: \\
WLOP's new art collection Ghostblade 4 is out!! Here we're giving away a brand new artbook to our lovely friends! \textbf{You'll need at least 80 likes and 10 comments to participate}. The lottery draw will be held at 7 PM on August 29! Everyone, don't forget to like, comment, and share! Features newly drawn illustrations from the Ghostblade and Cloud Insect series created between 2022 and 2025 ……

\textbf{Comments}:\\
“So stunning! [Sparkling eyes] And please pick me for the giveaway! [Grin] [Smirking]”

“I’ve always adored WLOP’s art style [Cloud Zoo Hot Air Balloon]. Hearing you say the Ghostblade 4 collector’s edition has amazing quality and is super-sized has totally raised my expectations. I really want to feel that texture in my own hands!”

“Wow, this is so exquisite!” ……

“Love it! [Cheering]”\\

Example2\\

\textbf{Title}: \\
Have You Slept Well, Puppy?

\textbf{Tag}: \\
Sleep, Sleep aid, Hypnosis, Livestream, Streamer

\textbf{ASR}: \\
(BGM) Wanna be in it forever girl, you know I adore ya ……

\textbf{Comments}:\\
“I love this angle, it's perfect! [heart]”

“Wow wow wow [cheering][cheering][cheering]. I can't fall asleep now! ”

“Love it [heart]”

“I really like this one.”

“I like you [giving flowers]”

\end{tcolorbox}
\caption{Representative examples of “inflated bubbles”: videos with high popularity metrics that experts rated as low-quality. Although user comments all show positive sentiment, experts have determined that these videos contain giveaway incentives and guided commenting behaviors, or include suggestive content, and therefore believe the video quality should be rated as low-quality.}
\label{fig:cases}
\end{figure*}

\begin{figure*}[t]
\centering
\begin{tcolorbox}[
  width=\textwidth,
  colback=white,
  colframe=black,
  title=Oracle Social Context: CoT generated by Gemini conditioned on real user comments,
  listing only,
  listing options={
    basicstyle=\small\ttfamily,
    breaklines=true
  }
]
\small
Based on the provided video metadata, this video belongs to the ``Travel \& Outdoors'' primary category and the ``Travel Vlog'' subcategory. The title, ``Studying abroad or going overseas cannot change my life,'' directly highlights the core reflection of the video. The video duration is ``288 seconds,'' with a resolution of ``3840$\times$2160,'' and it is not in vertical format, indicating relatively high visual quality. The ASR content reveals deeper philosophical reflections, covering travel experiences in ``Iceland,'' a redefinition of the concept of the ``wilderness,'' discussions on personal ``choice and responsibility,'' and references to and interpretations of the ``Frame 25'' metaphor from the film \emph{The Secret Life of Walter Mitty}, ultimately emphasizing the importance of ``determination.'' The cover image text, ``The wilderness is merely a determination,'' strongly aligns with the title and ASR content, jointly establishing a contemplative tone. Key frames showcase Icelandic natural landscapes, urban scenery, road trips, and moments of personal reflection by the creator, resulting in visually appealing and narratively rich visuals.

............

When viewers encounter the ASR content describing the creator's personal experiences and reflections on ``studying abroad,'' as well as discussions of ``courage'' and ``insight,'' audiences with similar life experiences tend to develop a strong sense of identification and resonance.
\textbf{[``I am a 25-fall freshman at a university in Singapore. This video gave me a lot of reflection, insight, and courage. Loved it [applause]'']} (supportive comment 1)

When viewers resonate with the ASR's in-depth discussion of ``travel'' and the delicate portrayal of the ``withdrawal'' state after returning from a trip, especially those who have just come back from traveling, they are deeply moved by how accurately the video captures their inner feelings.
\textbf{[[cry][cry][cry] I cried watching this. I just returned from Southern Europe yesterday and have been stuck in a withdrawal state. I happened to see this video on my feed. Thank you so much—it spoke directly to my heart.]} (supportive comment 2)

The ASR content deeply explores the core arguments of ``the wilderness as determination'' and ``taking responsibility for one's choices,'' prompting some viewers to engage in profound philosophical reflection and extend the video's viewpoints through their own experiences.
\textbf{[Long philosophical reflection elaborating on anxiety, responsibility, action, freedom, and the meaning of ``the wilderness,'' concluding that making a firm decision to confront life’s difficulties places one already within the wilderness.]} (supportive comment 3)

When viewers encounter the ASR's explicit statement that ``we are not free when we cannot take responsibility for what lies ahead or for our own choices,'' along with its interpretation of ``anxiety,'' they feel understood and emotionally healed.
\textbf{[Thank you, uploader! I keep rewatching this, and I really feel healed while being anxious [cry][cry]!! When I cannot take responsibility for what is in front of me or for my own choices, I am not free…]} (supportive comment 4)

............

When viewers encounter this philosophically rich ``travel vlog'' during their own travels, the alignment between the content and their current situation creates especially strong resonance.
\textbf{[I saw this video while living in Dali during my travels, and it really moved me.]} (supportive comment 14)

Regarding the discussion of ``freedom'' and ``value'' implied by the title ``Studying abroad or going overseas cannot change my life,'' some viewers adopt a critical stance, believing the creator failed to extract higher value from their experiences.
\textbf{[Your freedom did not bring you greater value.]} (opposing comment 1)

When the cover and ASR present ``the wilderness as determination'' as the core idea, some viewers perceive this insight as unoriginal or superficial and respond with sarcasm.
\textbf{[That just means you haven't matured mentally. If the wilderness is determination, then I already had it when I was six or seven.]} (opposing comment 2)

Given the high resolution, refined key frames, and depth of thought expressed in the ASR, viewers feel the video's high production quality deserves more attention and lament its relatively low view count.
\textbf{[This video is amazing! How does it only have 37k followers?!]} (supportive comment 15)

The ASR's final reveal of ``Frame 25'' and the creator's promise for the future provide profound insight and emotional elevation, leading viewers to regard the ending as exceptionally well-crafted.
\textbf{[The ending is brilliant.]} (supportive comment 16)

The ASR's deep interpretation of ``what the wilderness is'' and ``paying the price for one's actions'' offers strong emotional support and meaningful guidance to viewers facing life difficulties.
\textbf{[I really love this so much. You have no idea how important this was to me while I was going through a hard time. Especially the explanation of what the wilderness means and taking responsibility for one's actions—it strengthened my sense of meaning.]} (supportive comment 17)

In the simulated comments, there are \textbf{17 supportive comments} and \textbf{2 opposing comments}.
The Sigma-normalized difference (Skellam z-score) is computed as 
\[
z = \frac{17 - 2}{\sqrt{17 + 2}} = \frac{15}{\sqrt{19}} \approx 3.441.
\]
Since $z \geq 1.5$, the overall stance is classified as \textbf{supportive}.
\end{tcolorbox}
\caption{Oracle Social Context: Social-CoT reasoning path generated by Gemini, grounded in real high-engagement user comments. This represents the gold standard for social reasoning.}
\label{fig:cot_gemini}
\end{figure*}

\begin{figure*}[t]
\centering
\begin{tcolorbox}[
  width=\textwidth,
  colback=white,
  colframe=black,
  title=Social-CoT generated by MEDEA with Social Alignment Reward,
  listing only,
  listing options={
    basicstyle=\small\ttfamily,
    breaklines=true
  }
]
\small
Based on the provided script information, the video belongs to the \emph{Travel \& Outdoors} category, specifically the \emph{Travel Log} genre.
The title, \emph{``Iceland Vlog! One Day You Will Come to the End of the World,''} clearly establishes both the theme and emotional tone, framing the video as a documentary-style record of an Icelandic journey and emphasizing Iceland’s image as the ``end of the world.''

The video has a duration of \textbf{351 seconds}, a high resolution of \textbf{3840$\times$2160}, and is presented in a non-vertical format, indicating a high level of production quality with a strong emphasis on visual experience.
The ASR transcript provides a detailed narration of various locations in Iceland, including Reykjavík, the Snæfellsnes Peninsula, black sand beaches, and Vatnajökull Glacier.

............

When viewers observe that \emph{the title explicitly frames the content as an Iceland travel vlog} and that \emph{the ASR provides detailed descriptions of diverse locations and experiences}, they tend to appreciate the richness of the content.  
\textbf{[``Iceland is so beautiful, and the vlog is very rich in content.'']} (Supportive Comment~1)

Given the \emph{4K resolution (3840$\times$2160) and the keyframes displaying Iceland’s magnificent landscapes}, viewers are impressed by the visual clarity and image quality.  
\textbf{[``Iceland is stunning, and the visuals are incredibly sharp.'']} (Supportive Comment~2)

Based on the \emph{vivid ASR descriptions of Iceland’s scenery, such as the coexistence of an ``end-of-the-world'' atmosphere and ``soft pink gentleness''}, viewers develop a deeper understanding of Iceland’s unique charm.  
\textbf{[``I didn’t know Iceland was like this—it feels both apocalyptic and gently beautiful.'']} (Supportive Comment~3)

Considering the \emph{compelling presentation of Iceland throughout the video}, viewers express a desire to see more content about this destination.  
\textbf{[``So beautiful—please film more about Iceland.'']} (Supportive Comment~4)

From the \emph{creator’s insights and reflections conveyed in the ASR}, viewers perceive depth beyond surface-level scenery.  
\textbf{[``This is impressive—you really understand Iceland instead of just filming the scenery.'']} (Supportive Comment~5)

In response to the \emph{ASR’s description of extreme winds and keyframes showing people walking against strong gusts}, viewers are struck by Iceland’s raw natural power.  
\textbf{[``The wind in Iceland looks intense—really shocking.'']} (Supportive Comment~6)

Based on the \emph{shared travel experiences in the ASR}, viewers admire the creator’s extensive travel background.  
\textbf{[``Amazing—it feels like you’ve traveled all over the world.'']} (Supportive Comment~7)

Some viewers, drawing on their \emph{knowledge of Iceland}, offer mild suggestions while remaining supportive overall.  
\textbf{[``This is great, though Iceland also has hot springs and auroras that weren’t fully covered.'']} (Supportive Comment~8)

Viewers respond positively to the \emph{ASR’s depiction of southern Iceland}, expressing particular appreciation for that region.  
\textbf{[``This is impressive—I especially love the scenery in southern Iceland.'']} (Supportive Comment~9)

Those with \emph{personal travel experience in Iceland} find the ASR’s descriptions highly accurate.  
\textbf{[``I went to Iceland last year, and the video describes it perfectly.'']} (Supportive Comment~10)

From the \emph{overall production quality}, viewers recognize the creator’s dedication and effort.  
\textbf{[``This video is extremely well-made—you can really feel the effort.'']} (Supportive Comment~11)

Some viewers supplement the \emph{ASR’s travel advice} with personal experiences, while maintaining a supportive stance.  
\textbf{[``I’ve been to Iceland too—it’s not the easiest place to travel, but the video gives good warnings and is very well done.'']} (Supportive Comment~12)

Based on the \emph{ASR’s emphasis on Iceland’s colors}, viewers praise the video’s visual expressiveness.  
\textbf{[``Iceland looks so colorful in this video—it’s beautifully shot.'']} (Supportive Comment~13)

A few viewers note that the \emph{portrayal of Iceland’s characteristics could be more comprehensive}, but still offer positive feedback.  
\textbf{[``It doesn’t fully capture everything about Iceland, but it’s still great.'']} (Supportive Comment~14)

Some viewers express \emph{personal disagreement with the ASR’s positive framing of Iceland}, yet acknowledge the high production quality.  
\textbf{[``I don’t think Iceland is that beautiful—too cold and barren—but the video itself is very well made.'']} (Supportive Comment~15)

Among the simulated comments, there are \textbf{15 supportive comments} and \textbf{0 opposing comments}.
The Sigma-normalized difference (Skellam $z$-score) is computed as:
\[
z = \frac{15 - 0}{\sqrt{15 + 0}} = \frac{15}{\sqrt{15}} \approx 3.873.
\]

Since $z \geq 1.5$, the overall tendency is \textbf{supportive}.

\end{tcolorbox}
\caption{Social-CoT with Alignment: Reasoning paths generated by MEDEA trained with Social Alignment Reward. The model displays Empathetic Depth, instantiating diverse personas and nuanced emotional reactions (e.g., ``apocalyptic and gently beautiful'').}
\label{fig:cot_medea}
\end{figure*}

\begin{figure*}[t]
\centering
\begin{tcolorbox}[
  width=\textwidth,
  colback=white,
  colframe=black,
  title=Social-CoT generated by MEDEA without Social Alignment Reward,
  listing only,
  listing options={
    basicstyle=\small\ttfamily,
    breaklines=true
  }
]
\small
Based on the provided script information, the video belongs to the \emph{Travel \& Outdoors} category and is classified as a \emph{Travel Log} vlog.
The title, \emph{``Iceland Vlog! One Day You Will Come to the End of the World,''} clearly establishes both the theme and emotional tone.
The video has a duration of \textbf{351 seconds}, a resolution of \textbf{3840$\times$2160}, and is presented in a non-vertical format.

............

When viewers observe \emph{Iceland’s distinctive natural landscapes and cultural scenes presented in the video, along with the ASR’s emphasis on rich colors}, they tend to be impressed by the visual expressiveness.  
\textbf{[``Beautiful—so full of life.'']} (Supportive Comment~1)

The \emph{overall presentation of Iceland’s scenery}, and the \emph{poetic descriptions of natural landscapes in the ASR}, easily evokes strong emotional impact and longing.  
\textbf{[``So beautiful!!! It makes me really want to go!'']} (Supportive Comment~2)

Given the \emph{high resolution (3840$\times$2160), refined keyframes, and vivid ASR descriptions}, viewers perceive the video as highly engaging and hope to see more similar travel vlogs in the future.  
\textbf{[``So beautiful—looking forward to more travel vlogs like this. Amazing.'']} (Supportive Comment~3)

The \emph{unique landforms and majestic natural scenery of Iceland}, together with the ASR’s framing of Iceland as the ``end of the world,'' provide strong visual enjoyment and trigger an immersive travel impulse.  
\textbf{[``So beautiful—I want to go to Iceland.'']} (Supportive Comment~4)

Through \emph{exquisite visuals and emotionally expressive narration in the ASR}, the video conveys Iceland’s mystery and grandeur, inspiring a strong desire to travel.  
\textbf{[``I really want to go—it's so beautiful.'']} (Supportive Comment~5)

The \emph{careful capture and artistic presentation of Iceland’s scenery} offer viewers deep visual satisfaction and a strong sense of resonance and longing.  
\textbf{[``So beautiful! I really want to go.'']} (Supportive Comment~6)

By presenting Iceland’s landscapes in \emph{high resolution} and pairing them with \emph{detailed ASR descriptions of natural wonders}, the video fosters viewers’ yearning for Iceland.  
\textbf{[``So beautiful—really want to go.'']} (Supportive Comment~7)

The \emph{distinctive and captivating scenery of Iceland}, together with shared travel experiences in the ASR, strongly moves viewers and fills them with anticipation.  
\textbf{[``I really want to go—it's so beautiful.'']} (Supportive Comment~8)

As a \emph{high-quality Iceland vlog}, the video’s depth of content and visual expressiveness lead viewers to rate its overall quality very highly and regard it as potentially classic.  
\textbf{[``A classic.'']} (Supportive Comment~9)

The \emph{unique natural landscapes of Iceland} and the ASR’s articulation of its distinctive charm stimulate viewers’ desire to travel, often expressed in a lighthearted manner.  
\textbf{[``So beautiful! I really want to go.'']} (Supportive Comment~10)

Through \emph{careful editing and philosophically reflective narration in the ASR}, the video not only showcases Iceland’s beauty but also encourages viewers to reflect on the meaning of travel itself.  
\textbf{[``I really want to go! Watching this makes me want to travel!'']} (Supportive Comment~11)

The \emph{vast and unique scenery of Iceland}, along with ASR descriptions of the coexistence of desolation and vitality, strongly attracts viewers and motivates exploration.  
\textbf{[``So beautiful!! I really want to go!!'']} (Supportive Comment~12)

By \emph{comprehensively presenting Iceland’s landscapes} and sharing immersive travel experiences through the ASR, the video sparks strong desire to experience Iceland firsthand.  
\textbf{[``So beautiful—I want to go.'']} (Supportive Comment~13)

The \emph{stunning scenery of Iceland}, combined with ASR descriptions of its unique atmosphere, deeply moves viewers and prompts genuine admiration and longing.  
\textbf{[``So beautiful—I really want to go.'']} (Supportive Comment~14)

With \emph{high-resolution visuals} and \emph{vivid ASR depictions of Iceland’s beauty}, the video pushes viewers’ desire to visit Iceland to its peak.  
\textbf{[``I really want to go! It's so beautiful!'']} (Supportive Comment~15)

The \emph{presentation of Iceland’s distinctive landscapes}, together with detailed ASR narration of the travel experience, provides both visual and emotional fulfillment, resulting in strong resonance.  
\textbf{[``So beautiful—I really want to go.'']} (Supportive Comment~16)

The \emph{magnificent natural scenery of Iceland}, paired with ASR descriptions of strong winds and black sand beaches, highlights Iceland’s uniqueness and further stimulates travel motivation.  
\textbf{[``I really want to go—it's so beautiful!'']} (Supportive Comment~17)

Among the simulated comments, there are \textbf{17 supportive comments} and \textbf{0 opposing comments}.
The Sigma-normalized difference (Skellam $z$-score) is computed as:
\[
z = \frac{17 - 0}{\sqrt{17 + 0}} = \frac{17}{\sqrt{17}} \approx 4.123.
\]

Since $z \geq 1.5$, the overall tendency is \textbf{supportive}.

\end{tcolorbox}
\caption{Social-CoT without Alignment: Reasoning paths generated by MEDEA trained without Social Alignment Reward. The output exhibits Social Mode Collapse, characterized by repetitive, robotic phrasing (``So beautiful'') lacking authentic community voice.}
\label{fig:cot_medea_no_real}
\end{figure*}

\begin{figure*}[t]
\centering
\begin{tcolorbox}[
  width=\textwidth,
  colback=white,
  colframe=black,
  title=Prompt used to generate reasoning content
]
\small

<Task Definition>

You are an expert in user-generated content analysis. Based on the provided visual features and textual
content of a video, you are required to first select 15--20 highly relevant comments from the
given set of top-liked comments that are strongly associated with the video content. Then, determine
the stance of each selected comment toward the video (supportive or opposing). Finally, generate
a reasoning process that simulates how viewers might comment on the video’s creative quality
based on its content, and summarize the overall stance of the comment section.

--------------------------------------------------

Input Data

1. Cover Image: The video’s cover image
2. Key Frames: Seven key frames extracted from the video
3. Title: \{title\}
4. Tags: \{tag\}
5. ASR: \{asr\}
6. Primary Category: \{new\_tid\_name\}
7. Secondary Category: \{new\_sub\_tid\_name\}
8. Duration: \{duration\}
9. Resolution: \{resolution\}
10. Vertical Format: \{vertical\}
11. Top-liked Comments: A pool of high-like comments from which 15--20
    strongly content-related comments must be selected

--------------------------------------------------

Output Requirements

The output must strictly follow JSON format:

\{
  "think": "<think>Natural and coherent inferred reasoning based on the selected comments.
  The reasoning must conclude with a statistical analysis and an overall stance judgment.</think>",
  "answer": "<answer>Support / Not Clearly Supportive</answer>"
\}

--------------------------------------------------

Comment Selection Rules

From the pool of top-liked comments, select 15--20 comments that are strongly associated
with the video content:

1. Exact Content Matching (Highest Priority):
   Comments should directly correspond to specific elements of the video content.
   Examples:
   - ``This looks amazing'' $\rightarrow$ linked to visual features
   - ``The mixed language makes it hard to understand'' $\rightarrow$ linked to ASR content

2. Thematic Relevance (Secondary Priority):
   Comments should relate to the overall theme or quality of the video.
   Examples:
   - ``The image quality is too blurry'' $\rightarrow$ linked to visual resolution
   - ``This is a waste of time'' $\rightarrow$ linked to perceived content value

3. Mandatory Exclusion Rule:
   Comments referring to auditory or sound-related elements must be excluded.

4. Handling Offensive Comments:
   Highly liked comments containing insults toward the uploader should be categorized as
   opposing the video’s creative quality and retained if they satisfy content relevance criteria.

--------------------------------------------------

Reasoning Process Construction Rules

1. Independent Coverage Requirement:
   Each selected comment must appear at least once independently.
   Merging or collapsing similar comments is prohibited.

2. Video--Comment Alignment:
   - Precise alignment:
     ``When viewers see \{visual information\} / read \{ASR content\},
     they may express \{comment\}.''
   - Thematic alignment:
     ``Given the video’s overall characteristics, it may lead to comments such as \{comment\}.''
   Only the provided 11 video attributes may be referenced.

3. Speculative Expression Style:
   Use inferential phrasing such as ``viewers may point out...'' and incorporate audience expectations.

4. Mandatory Statistical Summary:
   - Report the number of supportive and opposing comments.
   - Ensure strict numerical consistency.
   - Compute the Sigma-normalized difference (Skellam z-score):
     z = (X - Y) / sqrt(X + Y)
   - Decision rule:
     If z $\geq$ 1.5, conclude Support; otherwise, Not Clearly Supportive.
   - The z-score must be enclosed in boxed\{\}.

--------------------------------------------------

Overall Stance Determination

The overall stance is determined solely based on the simulated comments and the computed
Skellam z-score.

--------------------------------------------------

Reasoning Format Requirements

1. Insert a blank line between each simulated comment.
2. Use \texttt{<video>} to mark video information and \texttt{<comment>} to mark simulated comments.
3. Annotate each comment with its stance and index:
   - Support Comment + index
   - Opposing Comment + index

--------------------------------------------------

<Current Task>

Cover Image: <image>
Key Frames: <image><image><image><image><image><image><image>
Title: \{video\_title\}
Tags: \{video\_tag\}
ASR: \{video\_asr\}
Primary Category: \{video\_new\_tid\_name\}
Secondary Category: \{video\_new\_sub\_tid\_name\}
Duration: \{video\_duration\}
Resolution: \{video\_resolution\}
Vertical Format: \{video\_vertical\}
Top-liked Comments: \{video\_comments\}

Please strictly output the result in JSON format and do not include any additional explanations.
\end{tcolorbox}
\caption{Prompt used to generate reasoning content.}
\label{fig:gemini}
\end{figure*}

\begin{figure*}[t]
\centering
\begin{tcolorbox}[
  width=\textwidth,
  colback=white,
  colframe=black,
  title=Prompt used to train MEDEA,
  listing only,
  listing options={
    basicstyle=\small\ttfamily,
    breaklines=true
  }
]
\small

<Task Definition>

You are an expert in user-generated content analysis.
Given the visual characteristics and textual information of a video, you are required to simulate the types of comments that may appear in the comment section regarding the *creation quality* of the video.
Generate at least 15 distinct comments with clear stances, and finally determine the overall tendency of the comment section.

Input Data

1. Cover Image: The video’s cover image

2. Key Frames: Seven key frames extracted from the video

3. Title: \{title\}

4. Tags: \{tag\}

5. ASR: \{asr\}

6. Primary Category: \{new\_tid\_name\}

7. Secondary Category: \{new\_sub\_tid\_name\}

8. Duration: \{duration\}

9. Resolution: \{resolution\}

10. Vertical Format: \{vertical\}

Criteria for Overall Comment Tendency

1. The simulated comments must contain at least 15 entries.
   All comments must be non-duplicated and explicitly appear in the reasoning process.
   
2. Assume that among the simulated comments:

   - X comments are classified as *supportive*
   
   - Y comments are classified as *opposing*
   
3. Compute the Sigma-normalized difference (Skellam z-score):

   z = (X - Y) / sqrt(X + Y)
   
4. If z $\geq$ 1.5, the overall comment tendency is classified as "Support";
   otherwise, it is classified as "Not Clearly Supportive".
   
5. In the output, the z value must be wrapped using boxed{{}},
   for example: "z = boxed{{-2}}".
   
6. The numbers of supportive and opposing comments reported in the final summary must strictly match those generated during the reasoning process. Fabrication or inconsistency is not allowed.

<Current Task>

Cover Image: <image>

Key Frames: <image><image><image><image><image><image><image>

Title: 

Tags: 

ASR: 

Primary Category: 

Secondary Category: 

Duration: 

Resolution: 

Vertical Video: 

Please directly output the final result ("Support" or "Not Clearly Supportive") without providing any additional explanation.
\end{tcolorbox}
\caption{Prompt used to train MEDEA.}
\label{fig:medea}
\end{figure*}

\section{Detailed Results of Baselines}
\label{app:detailed_results}

Most of the compared baselines are originally designed as regression-based methods, which output continuous quality scores rather than discrete class labels. To ensure a fair and informative comparison under the classification setting adopted in this work, we perform threshold sweeping on the CASTER-Bench for all regression-based methods.

Specifically, for each method, we vary the decision threshold that maps predicted quality scores to discrete quality categories and evaluate the corresponding classification performance. The threshold that yields the best macro-averaged F1 score is selected and reported as the main result in the paper. This procedure allows each method to operate under its optimal decision boundary, avoiding performance degradation caused by suboptimal or arbitrary threshold choices.

We present the complete performance results of each method under different threshold settings. Detailed results for FastVQA, DOVER, MaxVQA, Q-Align, FineVQ, and VQA2 can be found in Table~\ref{tab:fastvqa_threshold}, Table~\ref{tab:dover_threshold}, Table~\ref{tab:maxvqa_threshold}, Table~\ref{tab:q-align_threshold}, Table~\ref{tab:finevq_threshold}, and Table~\ref{tab:vqa2_threshold}, respectively.
\begin{table*}[t]
\centering
\small
\setlength{\tabcolsep}{5pt}
\begin{tabular}{lcccccccccc}
\toprule
\multirow{2}{*}{threshold} & \multicolumn{3}{c}{High-Quality} & \multicolumn{3}{c}{Low-Quality} & \multicolumn{3}{c}{Macro Average} \\
\cmidrule(lr){2-4} \cmidrule(lr){5-7} \cmidrule(lr){8-10}
& Precision & Recall & F1 & Precision & Recall & F1 & Precision & Recall & F1 \\
\midrule
0.055 & 0.277 & \textbf{1.000} & 0.434 & 0.000 & 0.000 & 0.000 & 0.138 & 0.500 & 0.217 \\
0.065 & 0.277 & \textbf{1.000} & 0.434 & \textbf{1.000} & 0.001 & 0.002 & \textbf{0.639} & 0.500 & 0.218 \\
0.275 & 0.277 & 0.966 & 0.431 & 0.725 & 0.034 & 0.066 & 0.501 & 0.500 & 0.248 \\
0.485 & 0.292 & 0.903 & \textbf{0.441} & 0.812 & 0.161 & 0.269 & 0.552 & 0.532 & 0.355 \\
0.695 & 0.326 & 0.676 & 0.440 & 0.790 & 0.465 & 0.585 & 0.558 & \textbf{0.571} & 0.513 \\
0.795$^\star$ & 0.347 & 0.440 & 0.388 & 0.761 & 0.682 & 0.719 & 0.554 & 0.561 & \textbf{0.554} \\
0.915 & \textbf{0.500} & 0.002 & 0.005 & 0.723 & 0.999 & \textbf{0.839} & 0.612 & 0.501 & 0.422 \\
0.925 & 0.000 & 0.000 & 0.000 & 0.723 & \textbf{1.000} & \textbf{0.839} & 0.362 & 0.500 & 0.420 \\
\bottomrule
\end{tabular}
\caption{Performance comparison using FastVQA under different thresholds. Best threshold is marked with $\star$, and best results in each column are highlighted in bold.}
\label{tab:fastvqa_threshold}
\end{table*}

\begin{table*}[t]
\centering
\small
\setlength{\tabcolsep}{5pt}
\begin{tabular}{lcccccccccc}
\toprule
\multirow{2}{*}{threshold} & \multicolumn{3}{c}{High-Quality} & \multicolumn{3}{c}{Low-Quality} & \multicolumn{3}{c}{Macro Average} \\
\cmidrule(lr){2-4} \cmidrule(lr){5-7} \cmidrule(lr){8-10}
& Precision & Recall & F1 & Precision & Recall & F1 & Precision & Recall & F1 \\
\midrule
3.226 & 0.277 & \textbf{1.000} & 0.434 & 0.000 & 0.000 & 0.000 & 0.138 & 0.500 & 0.217 \\
4.226 & 0.277 & \textbf{1.000} & 0.434 & \textbf{1.000} & 0.001 & 0.002 & \textbf{0.639} & 0.500 & 0.218 \\
24.226 & 0.293 & 0.951 & \textbf{0.448} & 0.865 & 0.119 & 0.210 & 0.579 & 0.535 & 0.329 \\
44.226 & \textbf{0.309} & 0.606 & 0.409 & 0.761 & 0.480 & 0.589 & 0.535 & \textbf{0.543} & 0.499 \\
54.226$^\star$ & 0.308 & 0.377 & 0.339 & 0.739 & 0.676 & 0.706 & 0.524 & 0.526 & \textbf{0.523} \\
65.226 & 0.250 & 0.095 & 0.138 & 0.720 & 0.891 & 0.796 & 0.485 & 0.493 & 0.467 \\
85.226 & 0.000 & 0.000 & 0.000 & 0.723 & 0.999 & \textbf{0.839} & 0.361 & 0.500 & 0.419 \\
86.226 & 0.000 & 0.000 & 0.000 & 0.723 & \textbf{1.000} & \textbf{0.839} & 0.362 & 0.500 & 0.420 \\
\bottomrule
\end{tabular}
\caption{Performance comparison using DOVER under different thresholds. Best threshold is marked with $\star$, and best results in each column are highlighted in bold.}
\label{tab:dover_threshold}
\end{table*}

\begin{table*}[t]
\centering
\small
\setlength{\tabcolsep}{5pt}
\begin{tabular}{lcccccccccc}
\toprule
\multirow{2}{*}{threshold} & \multicolumn{3}{c}{High-Quality} & \multicolumn{3}{c}{Low-Quality} & \multicolumn{3}{c}{Macro Average} \\
\cmidrule(lr){2-4} \cmidrule(lr){5-7} \cmidrule(lr){8-10}
& Precision & Recall & F1 & Precision & Recall & F1 & Precision & Recall & F1 \\
\midrule
24.239 & 0.277 & \textbf{1.000} & 0.434 & 0.000 & 0.000 & 0.000 & 0.138 & 0.500 & 0.217 \\
25.239 & 0.276 & 0.998 & 0.433 & 0.000 & 0.000 & 0.000 & 0.138 & 0.499 & 0.216 \\
41.239 & 0.280 & 0.971 & 0.434 & 0.793 & 0.043 & 0.081 & 0.536 & 0.507 & 0.258 \\
57.239 & 0.313 & 0.818 & \textbf{0.453} & \textbf{0.818} & 0.313 & 0.453 & \textbf{0.565} & 0.565 & 0.453 \\
67.239$^\star$ & 0.345 & 0.518 & 0.414 & 0.772 & 0.623 & 0.690 & 0.558 & \textbf{0.571} & \textbf{0.552} \\
73.239 & \textbf{0.352} & 0.316 & 0.333 & 0.748 & 0.777 & 0.762 & 0.550 & 0.547 & 0.548 \\
89.239 & 0.250 & 0.002 & 0.005 & 0.723 & 0.997 & 0.838 & 0.486 & 0.500 & 0.422 \\
90.239 & 0.000 & 0.000 & 0.000 & 0.723 & \textbf{1.000} & \textbf{0.839} & 0.362 & 0.500 & 0.420 \\
\bottomrule
\end{tabular}
\caption{Performance comparison using MaxVQA under different thresholds. Best threshold is marked with $\star$, and best results in each column are highlighted in bold.}
\label{tab:maxvqa_threshold}
\end{table*}


\begin{table*}[t]
\centering
\small
\setlength{\tabcolsep}{5pt}
\begin{tabular}{lcccccccccc}
\toprule
\multirow{2}{*}{threshold} & \multicolumn{3}{c}{High-Quality} & \multicolumn{3}{c}{Low-Quality} & \multicolumn{3}{c}{Macro Average} \\
\cmidrule(lr){2-4} \cmidrule(lr){5-7} \cmidrule(lr){8-10}
& Precision & Recall & F1 & Precision & Recall & F1 & Precision & Recall & F1 \\
\midrule
0.109 & 0.277 & \textbf{1.000} & 0.434 & 0.000 & 0.000 & 0.000 & 0.138 & 0.500 & 0.217 \\
0.119 & 0.277 & \textbf{1.000} & 0.434 & \textbf{1.000} & 0.001 & 0.002 & 0.639 & 0.500 & 0.218 \\
0.319 & 0.277 & 0.995 & 0.433 & 0.750 & 0.006 & 0.011 & 0.514 & 0.500 & 0.222 \\
0.529 & 0.288 & 0.944 & \textbf{0.442} & 0.835 & 0.108 & 0.191 & 0.562 & 0.526 & 0.317 \\
0.739 & 0.359 & 0.484 & 0.412 & 0.772 & 0.668 & 0.716 & 0.565 & 0.576 & 0.564 \\
0.759$^\star$ & 0.382 & 0.404 & 0.392 & 0.766 & 0.749 & 0.758 & 0.574 & \textbf{0.577} & \textbf{0.575} \\
0.949 & \textbf{1.000} & 0.002 & 0.005 & 0.724 & \textbf{1.000} & \textbf{0.840} & \textbf{0.862} & 0.501 & 0.422 \\
0.959 & 0.000 & 0.000 & 0.000 & 0.723 & \textbf{1.000} & 0.839 & 0.362 & 0.500 & 0.420 \\
\bottomrule
\end{tabular}
\caption{Performance comparison using Q-Align under different thresholds. Best threshold is marked with $\star$, and best results in each column are highlighted in bold.}
\label{tab:q-align_threshold}
\end{table*}

\begin{table*}[t]
\centering
\small
\setlength{\tabcolsep}{5pt}
\begin{tabular}{lcccccccccc}
\toprule
\multirow{2}{*}{threshold} & \multicolumn{3}{c}{High-Quality} & \multicolumn{3}{c}{Low-Quality} & \multicolumn{3}{c}{Macro Average} \\
\cmidrule(lr){2-4} \cmidrule(lr){5-7} \cmidrule(lr){8-10}
& Precision & Recall & F1 & Precision & Recall & F1 & Precision & Recall & F1 \\
\midrule
20.0 & 0.277 & \textbf{1.000} & 0.434 & 0.000 & 0.000 & 0.000 & 0.138 & 0.500 & 0.217 \\
21.0 & 0.277 & \textbf{1.000} & 0.434 & \textbf{1.000} & 0.001 & 0.002 & \textbf{0.639} & 0.500 & 0.218 \\
33.0 & 0.291 & 0.932 & \textbf{0.443} & 0.831 & 0.129 & 0.223 & 0.561 & 0.530 & 0.333 \\
46.0 & 0.297 & 0.582 & 0.393 & 0.747 & 0.473 & 0.580 & 0.522 & 0.527 & 0.486 \\
54.0$^\star$ & 0.323 & 0.343 & 0.333 & 0.742 & 0.724 & 0.733 & 0.532 & \textbf{0.534} & \textbf{0.533} \\
60.0 & 0.284 & 0.134 & 0.182 & 0.724 & 0.870 & 0.791 & 0.504 & 0.502 & 0.486 \\
72.0 & \textbf{0.333} & 0.005 & 0.010 & 0.723 & 0.996 & 0.838 & 0.528 & 0.501 & 0.424 \\
73.0 & 0.000 & 0.000 & 0.000 & 0.723 & \textbf{0.999} & \textbf{0.839} & 0.361 & 0.500 & 0.419 \\
\bottomrule
\end{tabular}
\caption{Performance comparison using FineVQ under different thresholds. Best threshold is marked with $\star$, and best results in each column are highlighted in bold.}
\label{tab:finevq_threshold}
\end{table*}

\begin{table*}[t]
\centering
\small
\setlength{\tabcolsep}{5pt}
\begin{tabular}{lcccccccccc}
\toprule
\multirow{2}{*}{threshold} & \multicolumn{3}{c}{High-Quality} & \multicolumn{3}{c}{Low-Quality} & \multicolumn{3}{c}{Macro Average} \\
\cmidrule(lr){2-4} \cmidrule(lr){5-7} \cmidrule(lr){8-10}
& Precision & Recall & F1 & Precision & Recall & F1 & Precision & Recall & F1 \\
\midrule
0.206 & 0.277 & \textbf{1.000} & 0.434 & 0.000 & 0.000 & 0.000 & 0.139 & 0.500 & 0.217 \\
0.216 & 0.279 & 0.998 & 0.436 & \textbf{0.923} & 0.011 & 0.022 & \textbf{0.601} & 0.504 & 0.229 \\
0.396 & 0.308 & 0.890 & \textbf{0.458} & 0.847 & 0.233 & 0.365 & 0.577 & 0.562 & 0.412 \\
0.586 & 0.333 & 0.527 & 0.408 & 0.767 & 0.596 & 0.671 & 0.550 & 0.561 & 0.539 \\
0.616$^\star$ & \textbf{0.358} & 0.454 & 0.400 & 0.766 & 0.688 & 0.725 & 0.562 & \textbf{0.571} & \textbf{0.562} \\
0.786 & 0.347 & 0.122 & 0.181 & 0.730 & 0.912 & 0.811 & 0.539 & 0.517 & 0.496 \\
0.966 & 0.200 & 0.002 & 0.005 & 0.723 & 0.996 & 0.838 & 0.461 & 0.499 & 0.421 \\
0.976 & 0.000 & 0.000 & 0.000 & 0.723 & \textbf{1.000} & \textbf{0.839} & 0.361 & 0.500 & 0.420 \\
\bottomrule
\end{tabular}
\caption{Performance comparison using VQA2 under different thresholds. Best threshold is marked with $\star$, and best results in each column are highlighted in bold.}
\label{tab:vqa2_threshold}
\end{table*}

\section{Declaration of AI Assistance}
We utilized Gemini to refine the wording and correct grammatical errors in the drafting of this paper. The authors reviewed and revised all AI-generated suggestions to ensure accuracy and consistency with the original ideas.

\end{document}